\pdfoutput=1

\documentclass[11pt]{article}

\usepackage[final]{acl}

\usepackage{times}
\usepackage{latexsym}
\usepackage{amsfonts}
\usepackage{tcolorbox}
\usepackage{tikz}

\usepackage[T1]{fontenc}

\usepackage[utf8]{inputenc}

\usepackage{microtype}

\usepackage{inconsolata}

\usepackage{graphicx}
\usepackage{multirow}
\usepackage{amsmath} 
\usepackage{booktabs}
\usepackage{pifont}%
\usepackage{subcaption}

\title{\emph{Less} is \emph{More}: Making Smaller Language Models  Competent Subgraph Retrievers for Multi-hop KGQA}

\author{
  Wenyu Huang$^{1}$,
  Guancheng Zhou$^{2}$,
  Hongru Wang$^{3}$, 
  Pavlos Vougiouklis$^{2}$, \\
  \textbf{Mirella Lapata$^{1}$, 
  Jeff Z. Pan$^{1,2}$\thanks{Corresponding author.}} \\
  $^1$School of Informatics, University of Edinburgh \\ 
  $^2$Huawei Edinburgh Research Centre, CSI, UK 
  $^3$The Chinese University of Hong Kong \\\texttt{w.huang@ed.ac.uk}, \texttt{guanchengzhou1@huawei.com}, \texttt{hrwang@se.cuhk.edu.hk}\\\texttt{pavlos.vougiouklis@huawei.com}, \texttt{mlap@inf.ed.ac.uk}, \\\texttt{http://knowledge-representation.org/j.z.pan/}
}

\begin{document}
\maketitle
\begin{abstract}
Retrieval-Augmented Generation (RAG) is widely used to inject external non-parametric knowledge into large language models (LLMs). Recent works suggest that Knowledge Graphs (KGs) contain valuable external knowledge for LLMs. Retrieving information from KGs differs from extracting it from document sets. Most existing approaches seek to directly retrieve relevant subgraphs, thereby eliminating the need for extensive SPARQL annotations, traditionally required by semantic parsing methods. In this paper, we model the subgraph retrieval task as a conditional generation task handled by small language models. Specifically, we define a subgraph identifier as a sequence of relations, each represented as a special token stored in the language models. Our base generative subgraph retrieval model, consisting of only 220M parameters, achieves competitive retrieval performance compared to state-of-the-art models relying on 7B parameters, demonstrating that small language models are capable of performing the subgraph retrieval task. Furthermore, our largest 3B model, when plugged with an LLM reader, sets new SOTA end-to-end performance on both the WebQSP and CWQ benchmarks. Our model and data will be made available online: \url{https://github.com/hwy9855/GSR}.

\end{abstract}

\section{Introduction}
\begin{figure}
    \centering
    \includegraphics[width=0.48\textwidth]{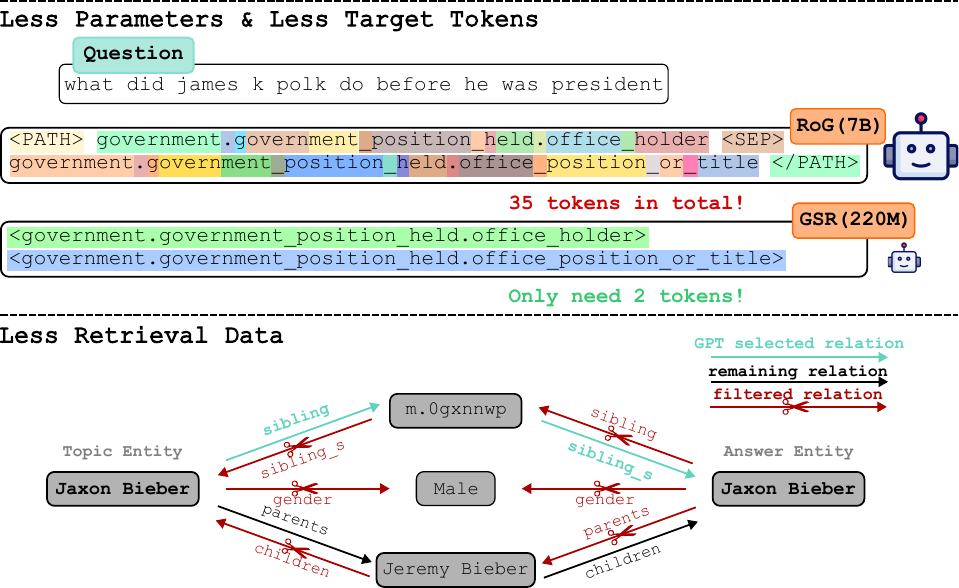}
    \caption{Our proposed GSR architecture facilitates training a smaller LM with \textbf{\emph{less}} parameters by reconsidering the subgraph retrieval task as a subgraph ID generation task, leading to shorter sequences of \textbf{\emph{less}} target tokens. In addition, we propose two ways of obtaining \textbf{\emph{less}} training samples from weakly supervised retrieval data to reduce the noise (e.g., path \textit{gender} -- \textit{gender} is not informative and brings noise in training).}
    \label{fig:motivation}
\end{figure}
Large Language Models (LLMs) have demonstrated tremendous capabilities in various Natural Language Processing tasks \citep{touvron2023llama, openai2024gpt4}. Despite their success, their hallucination tendencies still limit their performance across the involved tasks, and in question answering, in particular \citep{zhang2023sirens, huang2023survey}. Retrieval-augmented generation methods are widely used to enhance LLMs~\cite{PRKSC2023} and provide better safeguards against hallucination issues~\citep{shuster-etal-2021-retrieval-augmentation, tonmoy2024comprehensive}. Knowledge Graphs (KGs)~\cite{PVGW2017,PCEH+2017} have been recognised as valuable knowledge sources due to their compact triple representation, clear and noise-free knowledge format, and rich domain-specific information \citep{baek-etal-2023-knowledge, huang2024prompting}. Consequently, they have attracted significant attention by researchers who seek to propose efficient solutions that leverage information enclosed within a KG for question answering (KGQA) \citep{WHHH+2024,baek-etal-2023-knowledge, 10387715, DBLP:conf/aaai/0160LRSZD24}.

Traditional Semantic Parsing (SP) based KGQA \citep{das-etal-2021-case, hu-etal-2022-logical, luo2023chatkbqa} initiate the process by generating a SPARQL query derived from the natural language question. This query is subsequently executed against the KG of interest. Recently, Subgraph Retrieval (SR) based KGQA becomes popular as they eliminate the need of extensive SPARQL annotations (i.e. demonstrations consisting of (question, SPARQL) pairs). Typical SR-based KGQA methods \citep{zhang-etal-2022-subgraph, luo2024reasoning, sun2024thinkongraph} operate within a retrieval-augmented generation framework. They first retrieve a subgraph from the knowledge graph, which is then sent to a reader to generate answers. 
These methods achieve competitive performance compared with complex SP-based methods.

Since state-of-the-art (SOTA) methods for SR-based KGQA, such as those by \citet{luo2024reasoning} and \citet{sun2024thinkongraph}, typically employ LLMs to generate relevant relation chains,  they can be expensive in both training and reasoning due to the complexity of relation names (e.g., \textit{government.\-government\_position\_held.office\_position\_or\_title} is tokenized into 18 tokens, with the LLaMA2 tokenizer). In addition, the complexity of relation names induces an unnecessary task in mapping relations to relation names, besides mapping questions to relation chains. 
These drawbacks inspire us to consider a generative subgraph retriever architecture specified with smaller language models that have \textbf{\emph{less}} parameters. This forms the primary research question of this work: \textbf{How can small language models be utilized to accomplish the subgraph retrieval task with better efficiency and comparable or superior effectiveness compared to large language models?}

To simplify the task of generating relation chain given the complex relation names and to improve the representation of relations in language models, we treat each relation as a special token (relation ID) in the small language model. This transforms the subgraph retrieval task into predicting a sequence of relation IDs, which we define as the identifier of a subgraph (subgraph ID). As demonstrated in Figure~\ref{fig:motivation}, our redefined task requires much \textbf{\emph{less}} tokens (2 tokens vs. 35 tokens) as target, makes the training easier and increases the inference efficiency.
Our subgraph retriever, named Generative Subgraph Retriever (GSR), is trained jointly using two types of data: indexing data and subgraph retrieval data. In addition, we observe a large amount of noise inside the subgraph retrieval data used by most previous works \citep{zhang-etal-2022-subgraph, luo2024reasoning}. To mitigate this issue, we proposed two data pruning methods to obtain denoised retreival data with \textbf{\emph{less}} training samples.

Our comprehensive experiment showcase that employing large language model for (generative) sub-graph retrieval is an unnecessary expense. Our best setting, even when a 220M parameters model is used (30$\times$ \textbf{\emph{less}} parameters), can achieve +9.2\% and +5.3\% F1 score improvement (\textbf{\emph{more}} effective) on the WebQSP and CWQ benchmarks respectively over the previous SOTA subgraph retrieval work \citep{luo2024reasoning} consisting of 7B parameters. By integrating with our LLM reader, which is fine-tuned from the same base model used in the SOTA approaches, we achieve F1 score improvements of +6.3\% on WebQSP and +4.9\% on CWQ in the end-to-end evaluation. Our 3B model with LLaMA3 reader reaches new SOTA performance among SR-based KGQA on both WebQSP and CWQ dataset with F1 80.1\% and 64.4\% respectively.

Our contributions can be summarised as follows: 1) We introduce GSR, a method utilizing small language models to accomplish the subgraph retrieval task. 2) We propose a training framework comprising an indexing step and a retrieval step for training GSR, including: a) an automatic method for collecting indexing data; b) two distinct methods to enhance the quality of the retrieval data. 3) Comprehensive experimental results demonstrate the effectiveness of our work. Our best model achieves an average improvement of +5.6\% in F1 score on two KGQA datasets compared to the previous SOTA SR-based KGQA models, while being 7.7 times more efficient during the subgraph retrieval step.

\section{Related Works}
Our method draws inspiration from works on Knowledge Graph Question Answering (KGQA) and Generative Retrieval.

\subsection{KGQA}
KGQA is the task of answering questions based on facts from a knowledge graph. In general, KGQA methods can be classified into two categories: Semantic Parsing (SP)-based and Subgraph Retrieval (SR)-based KGQA. 

\paragraph{SP-based KGQA}
SP-based KGQA methods are designed to transform a question into an executable logical query \citep{hu-etal-2022-logical, das-etal-2021-case, luo2023chatkbqa}, which can then be directly applied to a KG to retrieve answers. These methods are famous for their versatility in handling diverse complex questions \citep{DBLP:journals/aiopen/ZhangZKLHSCL23}. 
Despite the effectiveness, SP-based KGQA methods generally requires extensive SPARQL annotations from experts, which is expensive to obtain in practice \citep{zhang-etal-2022-subgraph}. In addition, if the generated SPARQL is not executable, no answer will be generated \citep{luo2024reasoning}.

\paragraph{SR-based KGQA}
SR-based KGQA methods, on the other hand, present a different methodology for handling the KGQA task with a retrieval-augmented generation framework, which first retrieves relevant KG subgraphs, then uses a subgraph reader to generate the final answer. \citet{baek-etal-2023-knowledge} treat the subgraph retrieval process in triple level, where each triple is textualized as a document. \citet{zhang-etal-2022-subgraph} model the subgraph retrieval in the relation level, using a dual-encoder to retrieve relevant relation.
\citet{luo2024reasoning} and \citet{sun2024thinkongraph} conduct the subgrpah retrieval as a reasoning task, using a large language model to generate the reasoning step for subgraph retrieval. However, it is inefficient to rely on LLMs in subgrpah retrieval, which we consider a simple task that can be handled by smaller LMs with specific design.
In addition to KG retrieval for RAG, knowledge graphs can be useful for passage based RAG in many different ways, such as extracting knowledge graph triples for selecting the most relevant passages~\cite{GSGYS2024} or by using KG patterns to train some LLMs for planning the retrieval~\cite{WCHY+2024}. 

\subsection{Generative Retrieval}
Generative retrieval is a new paradigm of information retrieval (IR), framing traditional IR into a sequence-to-sequence modelling task \cite{DBLP:conf/emnlp/Pradeep0GLZLM023}. This paradigm works by storing a search index inside the model's parameters instead of outside,  treating it as an external module. Differentiable Search Index (DSI) is a series of typical generative retrieval techniques \citep{DBLP:conf/nips/Tay00NBM000GSCM22, DBLP:journals/corr/abs-2206-10128, chen-etal-2023-understanding}.
Generative retrieval methods have shown potential to outperform dual-encoder-based methods, but they face challenges with respect to scaling to large numbers of documents since the parameters of involved can be limited. In this work, instead of assigning each subgraph a specific ID, we decompose the subgraph retrieval task as a sequence generation task, where an auto-regressive model is responsible for decoding a relation chain, as a sequence of unique relation IDs.

\section{Problem Statement}
A Knowledge Graph $\mathcal{KG}=\{(s, r, o) \|s, o\in\mathcal{E}, r\in\mathcal{R}\}$ is an RDF graph that consists of several $(s, r, o)$ triples, where $\mathcal{E}$ is the entity set and $\mathcal{R}$ is the relation set, and $s, r, o$ are instances of a subject entity, a relation, and an object entity respectively, for a triple $\in \mathcal{KG}$. Knowledge Graph Question Answering (KGQA) aims to figure out the answer set $\hat{\mathbf{A}}$ given the question $q$ and topic entity $e_{\text{t}}$. KGQA methods based on subgraph retrieval model the question answering task in a retrieval-augment generation framework by first retrieving relevant subgraph $\mathcal{SG}\subseteq\mathcal{KG}$ given the question $q$, and, subsequently, using a reader to generate the predicted answer set $\mathbf{A}$ given the question $q$ and the retrieved subgraph $\mathcal{SG}$.

\section{Methodology}

\begin{figure}[t]
    \centering
    \includegraphics[width=0.48\textwidth]{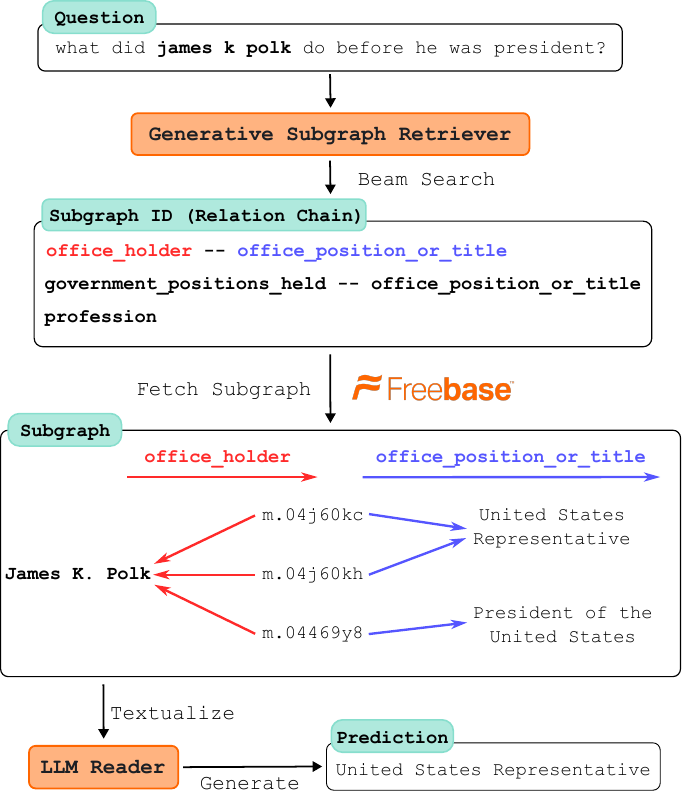}
    \caption{Overall pipeline of proposed RAG framework for KGQA task with the  Generative Subgraph Retriever.}
    \label{fig:gsr_arch}
\end{figure}

\subsection{Generative Subgraph Retriever}
\paragraph{Subgraph Definition}
\label{sec:subgraph}

A KG subgraph $\mathcal{SG}\subseteq\mathcal{KG}$ is a subset of all triples in the original KG. Specially, in this work, we made a further constraint on the definition of subgraph, which we called path constrained subgraph\footnote{In other part of the paper, we use the term `subgraph' to refer to path constrained subgraph if not specially mentioned.}. In short, path constrained subgraph is a set of special subgraph that can be identified with a simple identifier $(e, c)$, where $e$ is the entity and $c=r_1, r_2, ..., r_n$ is an ordered relation chain that indicates $n$-hop reasoning. By moving from the entity $e$ through the relation chain $c$, we can identify the whole subgraph with the simple identifier.
The middle part of Figure~\ref{fig:gsr_arch} shows an example of identify subgraph with topic entity \textit{James K. Polk} and relation chain (\textit{office\_holder}, \textit{office\_position\_or\_title}), which can be identified with identifier \texttt{[James K. Polk, office\_holder, office\_position\_or\_title]}. In addition, for better flexibility, we decouple the topic entity $e$ and relation chain $c$ in the subgraph identifier, which means that the two part can be retrieved separately. We assume the topic entity $e$ is known in this work for aligning other SR-based KGQA works \citep{zhang-etal-2022-subgraph, luo2024reasoning}.

\paragraph{Subgraph Retrieval}
With the subgraph definition above, we can model the subgraph retrieval task as a subgraph ID generation task. Given a natural language query $q$, subgraph retrieval aims to find the relevant subgraph, as a relation chain $c=\{r_1, r_2, ..., r_n\}$, that forms an $n$-hop reasoning chain from the topic entity $e_t$ to the answer entity $e_a$. The relation chain is ordered, and is predicted by the probability of 
\begin{equation*}
    P(c|q) = \Pi_i^nP(r_i|q, r_1, r_2, ..., r_{i-1}).
    \label{equ:subgraph_retrieval}
\end{equation*}
The above equation can be decomposed into the conventional auto-regressive language modelling objective, assuming relations are mapped into the involved model's token space.

\paragraph{Subgraph ID}
Since the major task of our GSR model is to map questions to a subgraph ID (relation chain), it is essential to build an efficient and effective way of representing the relations in the model. \citet{luo2024reasoning} simply ask LLMs to generate the whole name of the relation (e.g., \textit{tv.regular\_tv\_appearance.actor}), but it is not efficient and hard for LMs to learn the mapping. Instead, we adapt the 
atomic document representation methods from DSI-based works \citep{DBLP:conf/nips/Tay00NBM000GSCM22}. On relation level, we assign each relation an atomic ID, i.e., each relation is mapped to a special token in the generative language model. On subgraph level, we build a hierarchical indexing, i.e., each subgraph ID is mapped to the relation tokens which forms the reasoning chain. 

\subsection{Training GSR}

\label{sec:train}
To train the GSR model, we adapt a multi-task training setting with indexing data and (subgraph) retrieval data.
\paragraph{Indexing Data}
Unlike textual retrieval, in the subgraph retrieval task, the information that required to be retrieved is a sequence of relations. Thus instead of simply defining the indexing task as a relation name to relation ID mapping task, we build a question to relation ID task to teach language model how different relations (i.e. relation IDs) can be expressed in natural language questions. For each relation in Freebase, we first filter out extremely infrequent relations that do not have at least one triple available in the Freebase full dump\footnote{\url{https://developers.google.com/freebase}}. After that, we use the prompt provided in Appendix~\ref{app:prompt} to prompt GPT-4 for getting 10 question templates $t_{r_i}^{(j)}$ for each relation $r_i$. Finally, for each template, we randomly sample triples $(s, r_i, o)$ and use $s$ to replace the placeholder in the template. By far, we can get the indexing data with every valid relation have at most 10 diverse pseudo question. The task for a language model is to map these pseudo questions to the relation ID.
\paragraph{Retrieval Data}
Though obtaining annotated reasoning-based KGQA training data is easier than getting the expert annotated semantic-parsing based data, it is still hard to get the gold relation chain annotation. Previous works \citep{zhang-etal-2022-subgraph, luo2024reasoning} try to mitigate this issue by seeking the weakly supervised data constructed from the question answer pair. Given a question, raw weakly supervised data is collected from retrieving the shortest path between the topic entity and the target answer entity. 

\begin{figure}[t]
    \centering
    \includegraphics[width=1.0\linewidth]{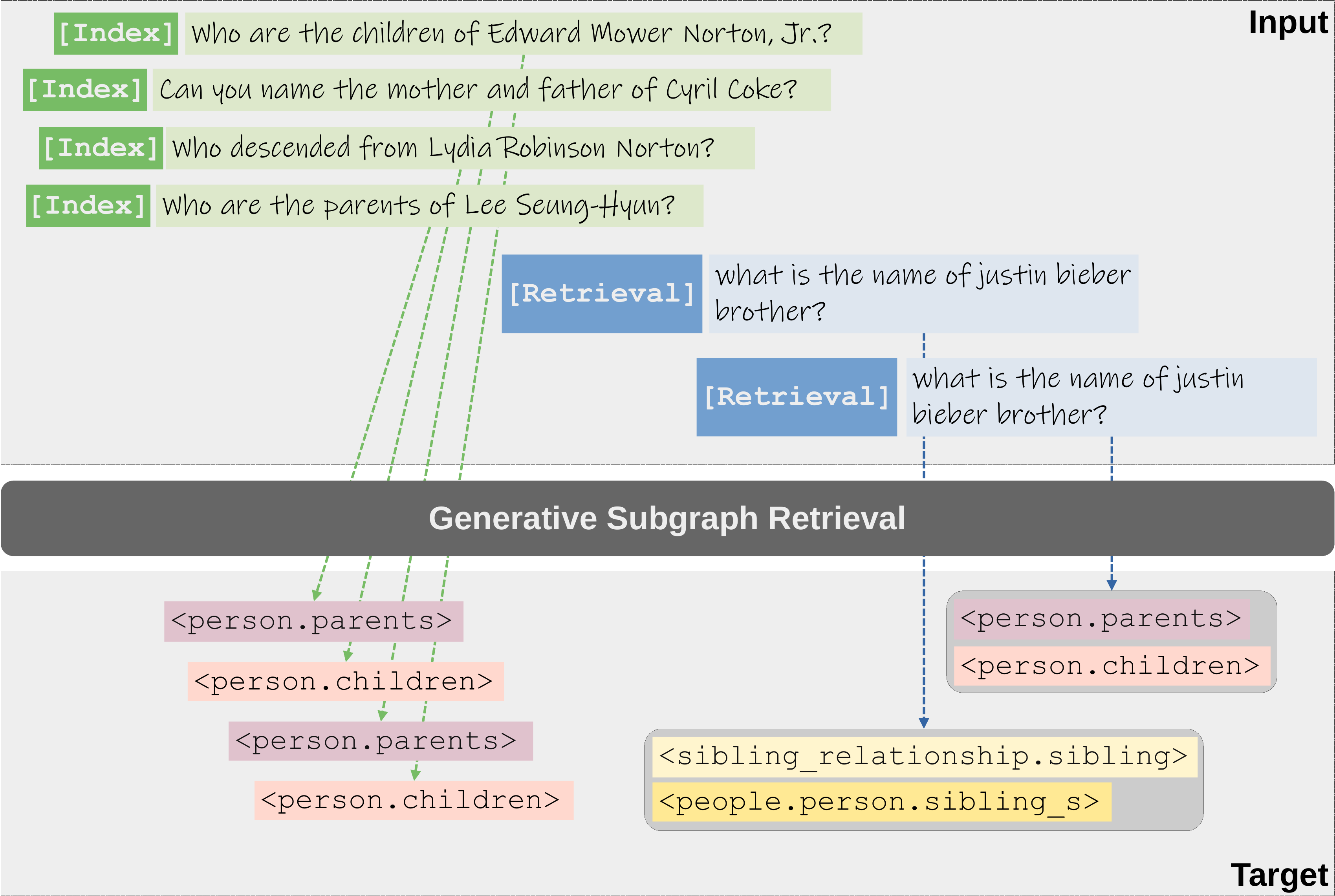}
    \caption{A demonstration of how we train the GSR model. Each coloured box stand for a special token in our GSR model. The indexing task is N to 1, where multiple pseudo questions is mapping to a specific relation ID. The subgraph retrieval task is 1 to N, where each question can be mapped to several subgraph IDs (sequence of relation IDs). 
    }
    \label{fig:gsr_model}
\end{figure}
However, we found the above data creation method results in a large amount of noise where the shortest path is meaningless. For example, when the topic entity and answer entity are both male person, then there must exist a path \textit{topic entity} $\rightarrow$ \textit{gender} $\rightarrow$ \textit{Male} $\leftarrow$ \textit{gender} $\leftarrow$ \textit{answer entity}. Training subgraph retrieval model with the raw data will cause model to generate useless chains, which can be harmful for the model performance. To address this, we only keep the shortest path that always has one direction that starts from the topic entity and ends at the answer entity and construct filtered retreival data. We observed that it is safe to do that since most of the relations also have a inverse relation in the Freebase so that we will not lose training signals by filtering out useful information. 

To get better training signal for training the retriever model, we further distil knowledge from GPT to get higher quality data. In short, we prompt GPT-4 with all shortest paths we get from the raw data and ask GPT-4 to choose the list the relevant relation chain with the prompt provided in Appendix~\ref{app:prompt}, and construct the GPT-selected retrieval data accordingly. By doing so we can both get rid of most noise training signal from raw data and avoid losing valuable training signal that is filtered in filtered data.

\paragraph{Training Strategy}
We adapt an jointly training strategy to train the GSR model, where the indexing data and the retrieval data is used at the same time for training the GSR model. Since we need to train the GSR model with both indexing data and retrieval data, we use a common strategy to distinguish the two task in the model level. Specifically, we design two special prefix token in front of the two different task, \texttt{[Index]} indicating this is a indexing task and \texttt{[Retrieval]} indicating this is a subgraph retrieval task. Figure~\ref{fig:gsr_model} demonstrate how we train the GSR model.
\subsection{Inference GSR}
When inference using the trained GSR model for subgraph retrieval task, we utilize beam search to retrieve top-$k$ subgraph ID (i.e., the relation chain).
In case a predicted relation chain is not executable against the knowledge graph (i.e. that we can not arrive to the end of the predicted chain starting from the topic entity of interest), it is removed from the set of available candidates. We retain top-$n$ valid relation chains where $n\leq k$ for balancing precision and recall of the subgraph retrieval step.

\subsection{LLM reader}

Inspired by \citet{luo2024reasoning}, we fine-tune an LLM as a reader for generating answer(s) based on the retrieved subgraph. We explore two approaches for injecting subgraph information into LLMs by representing subgraphs as reasoning paths or triple sets.

\paragraph{Subgraph as Reasoning Paths}
We adapt the prompting strategy from \citet{luo2024reasoning} to prompt LLMs with the complete reasoning path that leads from the topic entity $e_{start}$ to the potential answer entity $e_{end}$, by including any intermediate entity and relations in the path: 
$$e_{start} \text{ -> } rel_{int}^{(1)} \text{ -> } e_{int}^{(1)} \cdots \text{ -> } rel_{int}^{(n)} \text{ -> } e_{end}\text{,}$$
where $n$ is the number of hops between $e_{start}$ and $e_{end}$.
In order to consider inverse relations where entities after the right-arrow do not find themselves in the object position, we introduce bidirectional paths, changing the directionality of the arrows, accordingly. For instance, we might have paths as follows\footnote{Appendix~\ref{app:case_studies} shows some examples of inverse relations.}:
$$e_{start} \text{ <- } rel_{int}^{(1)} \text{ <- } e_{int}^{(1)} \cdots \text{ -> } rel_{int}^{(n)} \text{ -> } e_{end}$$
\paragraph{Subgraph as Triple Sets}
Though \citet{luo2024reasoning} has shown the effectiveness of representing subgraph as relation paths, it is expensive to use such paths since some triples will be repeated multiple times. To alleviate this issue, we tried to use another way to represent subgraph, which is to feed LLMs all the triples that appears in the retrieved subgraph. To this end, the knowledge would be denser in the context, but it is also harder for LLMs to build connections between relevant triples since they might be far away. 

\section{Experimental Setup}

\paragraph{Datasets}
We use WebQSP \citep{yih-etal-2016-value} and CWQ \citep{talmor-berant-2018-web} datasets that are both constructed from Freebase for all the experiments. Following previous works that seek to increase the reasoning efficiency on Freebase, we choose to use the same simplified Freebase (only contains multi-hop triples of all topic entities appeared in WebQSP and CWQ)~\citep{DBLP:conf/wsdm/HeL0ZW21, zhang-etal-2022-subgraph, luo2024reasoning}. Specially, WebQSP is a relatively easier dataset as its questions require at most 2-hop reasoning from their topic entity. CWQ is more challenging as the questions involve up to 4-hop reasoning. Detailed statistics of both datasets are provided in Appendix~\ref{app:datasets}. 

\paragraph{Baselines}
We compare our proposed RAG with GSR pipeline with several baselines. For evaluating subgraph retrieval performance, we compare our methods with \textbf{RoG} \citep{luo2024reasoning}. For end-to-end KGQA comparison, we choose both SP-based approaches including \textbf{UniKGQA} \citep{jiang2023unikgqa}, \textbf{DecAF} \citep{yu2023decaf}, \textbf{ChatKBQA} \citep{luo2023chatkbqa} and SR-based approaches including \textbf{SR} \citep{zhang-etal-2022-subgraph}, \textbf{KAPING} \citep{baek-etal-2023-knowledge}, \textbf{RoG}  \citep{luo2024reasoning} and \textbf{ToG} \citep{sun2024thinkongraph}.

\paragraph{Metrics}
For evaluating subgraph retrieval, we choose Precision, Recall and F1 to measure the answer coverage of the retrieved subgraphs. For the end-to-end evaluation, we use F1, Hits@1 \citep{zhang-etal-2022-subgraph, luo2023chatkbqa}. RoG, in particular, measures Hits@1 in an uncommon way \citep{luo2024reasoning}. In order to align their results with ours, we further added a Hits metric, which treats all answers generated by LLM as the top-1 answer and measures the Hits@1 score accordingly. For the original Hits@1, we use the first answer generated by LLM.

\paragraph{Implementation Details}
We build our GSR model on top of T5 models with size base, large and 3b, which consist of 220M, 770M and 3B parameters respectively. For beam search, we retrieve top-10 result and keep at most 3 valid relation chains for end-to-end evaluation\footnote{Some relation chains are not executable on Freebase. This means that there is no reasoning path starting from the topic entity with the decoded relation chain. Detailed results under different beam size settings can be found in Appendix~\ref{app:beam}}. For subgraph retrieval evaluation, we set the beam size to 3 for a fair comparison against RoG. For the LLM reader, we fine-tuned LLaMA2-chat-7B and LLaMA3-instruct-8B with QLora \citep{dettmers2023qlora} using the unsloth library\footnote{\url{https://github.com/unslothai/unsloth}}. More implementation details can be found in Appendix~\ref{app:implementation}.

\section{Results}
We evaluate our approach across both sugraph retrieval and the end-to-end QA.
\subsection{Subgraph Retrieval Performance}
\begin{table}[t]
    \centering
    \small
    \begin{tabular}{l|p{0.5cm}p{0.5cm}p{0.6cm}|p{0.5cm}p{0.5cm}p{0.5cm}}
    \toprule
        \multirow{2}{*}{\bf{Model}} & \multicolumn{3}{c}{\bf{WebQSP}} & \multicolumn{3}{|c}{\bf{CWQ}} \\ 
        & P & R & F1 & P & R & F1 \\ \midrule
        \multicolumn{7}{c}{\textit{LLM based subgraph retrieval}} \\ \midrule 
        RoG & 46.90 & 79.85 & 49.56 & 18.88 & 67.89 & 22.26 \\
        \midrule
        \multicolumn{7}{c}{\textit{ours w/ raw data}} \\ \midrule
        GSR-base & 42.68 & 84.59 & 47.87 & 18.34 & 72.14 & 22.20 \\
        GSR-large & 42.19 & 85.16 & 47.11 & 18.29 & 72.88 & 22.19 \\
        GSR-3b & 44.18 & 85.73 & 49.03 & 19.05 & 72.67 & 22.79 \\ \midrule 
        \multicolumn{7}{c}{\textit{ours w/ filtered data}} \\ \midrule
        GSR-base & 48.85 & 85.75 & 54.64 & 21.45 & 72.66 & 26.09 \\
        GSR-large & 47.79 & 86.15 & 53.77 & 21.66 & 72.41 & 26.34 \\
        GSR-3b & 48.32 & \textbf{86.22} & 53.95 & 21.30 & 72.81 & 25.93 \\ 
        \midrule
        \multicolumn{7}{c}{\textit{ours w/ GPT selected data}} \\ \midrule
        GSR-base & 53.94 & 85.49 & 58.77 & \textbf{23.02} & 72.28 & 27.59 \\
        GSR-large & 53.67 & 84.91 & 58.23 & 22.68 & 72.09 & 27.19 \\
        GSR-3b & \textbf{54.46} & 85.63 & \textbf{58.97} & \textbf{23.02} & \textbf{73.00} & \textbf{27.60} \\
        \bottomrule
    \end{tabular}
    \caption{Subgraph retrieval results. We set beam size $k$=3 for fair comparison with baseline.}
    \label{tab:subgraph_ret_res}
\end{table}

\begin{table}[t]
    \centering
    \small
    \begin{tabular}{l|ccc}
    \toprule
        \textbf{Model} & \textbf{WebQSP} & \textbf{CWQ} & \textbf{Inf. Time} \\ \midrule
        RoG+T5-base & 80.9 & 70.4 & 912s + 1,887s \\
        GSR-base & 85.5 & 72.3 & 117s + 258s \\ 
    \bottomrule
    \end{tabular}
    \caption{Subgraph retrieval performance (Recall). \textbf{Inf. Time} is the inference time on WebQSP + CWQ.}
    \label{tab:rog-small}
\end{table}
Table~\ref{tab:subgraph_ret_res} shows results on subgraph retrieval.
\paragraph{\emph{less} parameter(s) is \emph{more}}Compared to RoG that performs relation chain retrieval with an LLM, the proposed GSR models achieve better performance across most metrics with less model parameters. This is observed even in the case of the smallest GSR-base models with approximate 30$\times$ less parameters than RoG, where we can obtain an average of +4.5\% more recalled answers in the retrieved subgraph when training with the same retrieval data (raw). In addition, we can observe that within the GSR variant trained on the same retrieval data, larger model generally works better. We note a few cases in which GSR-large works slightly worse than GSR-base (e.g., in ours w/ GPT-selected data). We attribute this to the more invalid paths that are generated by GSR-large, which can be mitigated by setting beam size $k=10$ (cf. Figure~\ref{tab:subgraph_ret_res_top10} in Appendix~\ref{app:top10}).

\paragraph{\emph{less} training samples is \emph{more}}Among GSR models trained with different retrieval data, the raw data with all the shortest paths as weakly supervised signal performs the worst according to F1. While the Recall score seems to be competitive with other models, the low precision indicates that some boost in recall is simply obtained by uninformative chains, such as \textit{gender} $\rightarrow$ \textit{gender}. We included some case studies about this issue in Appendix~\ref{app:case_studies}. GSR models trained with filtered data work better, with respect to Recall performance, since their variations trained with GPT selected data achieve highest Precision and F1 score, showcasing the best balance between finding the answer and reducing unrelated information. 
When it comes to the end-to-end performance, in the context of retrieval-augmented generation, both recall and precision are important, since the LLMs' content window is limited. Thus, we choose the GSR models trained with GPT selected data for the end-to-end evaluation in the next step.

\paragraph{\emph{less} target tokens is \emph{more}}
We further trained a T5 model with the same input and output settings as RoG \citep{luo2024reasoning} using GPT-selected retrieval data to explore the benefits of our apporach. Table~\ref{tab:rog-small} shows the subgraph retrieval performance and inference time when inferring these two base-sized models. It is evident from the results that our GSR architecture achieves a +3.3\% average improvement of recall while being 7.4 times more efficient during inference, demonstrating that incorporating less target tokens for the subgraph ID generation task brings in both effectiveness and efficiency.
\subsection{End-to-End Performance} 
\begin{table}[t]
    \centering
    \small
    \begin{tabular}{l|p{0.5cm}p{0.5cm}p{0.6cm}|p{0.5cm}p{0.5cm}p{0.5cm}}
    \toprule
        \multirow{2}{*}{\bf{Model}} & \multicolumn{3}{c}{\bf{WebQSP}} & \multicolumn{3}{|c}{\bf{CWQ}} \\ 
        & H@1 & Hits & F1 & H@1 & Hits & F1 \\ \midrule
        \multicolumn{7}{c}{\textit{SP-based KGQA baseline}} \\ \midrule
        UniKGQA & 77.2 & - & 72.2 & 51.2 & - & 49.4\\
        DeCAF & 82.1 & - & 78.8 & 70.4 & - & - \\
        ChatKBQA & 86.4 & - & \textbf{83.5} & \textbf{86.0} & - & \textbf{81.3} \\
        \midrule
        \multicolumn{7}{c}{\textit{SR-based KGQA baseline}} \\ 
        \midrule
        SR & 68.9 & - & 64.1 & 50.2 & - & 47.1 \\
        KAPING & - & 73.9 & - & - & - & - \\
        RoG & 80.0 & 85.7 & 70.8 & 57.8 & 62.6 & 56.2 \\
        ToG\textsubscript{ChatGPT} & 76.2 & - & - & 57.1 & - & -\\ 
        ToG\textsubscript{GPT-4} & 82.6 & - & - & 67.6 & - & -\\ \midrule
        
        \multicolumn{7}{c}{\textit{ours w/ LLaMA 2 7B (QLora)}} \\ \midrule
        \it{None} & 62.3 & 67.6 & 48.5 & 35.3 & 38.9 & 31.5 \\ 
        GSR-base & 85.7 & 88.0 & 76.7 & 63.4 & 67.2 & 60.1 \\
        GSR-large & 85.6 & 87.7 & 76.4 & 63.2 & 67.3 & 60.2 \\
        GSR-3b & 86.5 & 88.6 & 77.1 & 64.3 & 68.3 & 61.1 \\ 
        \midrule
        \multicolumn{7}{c}{\textit{ours w/ LLaMA 3 8B (QLora)}} \\ \midrule
        \it{None} & 65.3 & 70.3 & 51.9 & 38.1 & 41.8 & 34.3 \\
        GSR-base & 86.5 & 88.3 & 78.8 & 66.4 & 69.6 & 63.1 \\
        GSR-large & 87.0 & 88.7 & 78.9 & 66.4 & 69.6 & 63.5 \\
        GSR-3b & \textbf{87.8} & \textbf{89.6} & 80.1 & 67.5 & \textbf{71.1} & 64.4 \\
        \bottomrule
    \end{tabular}
\caption{End-to-end KGQA performance with LLM reader. The GSR models are trained by gpt selected weakly supervised data. H@1 stands for Hits@1. \textit{None} indicating that we use QLora finetuned LLM without any subgraph information for ablation studies.}
    \label{tab:end_to_end_res}
\end{table}

Table~\ref{tab:end_to_end_res} shows the end-to-end KGQA performance. Generally, on the WebQSP dataset, our best GSR-3b model with LLaMA 3 8B reader achieves the best performance among all SR-based methods. In particular, our best model achieves +5.2\% improvement for Hits@1 compared to ToG with GPT-4 and +9.3\% improvement for F1 compared to RoG. Even the GSR-base model with LLaMA 2 7B reader outperforms RoG which uses a finetuned LLaMA 2 7B model as both its retriever and reader, showing that the subgraph retrieval task can be effectively handled by small language models. Our systems' performance is also on par with, if not exceeds, SOTA SP-based methods. The best GSR system manages to achieve higher Hits@1 score, even though it is trained using only weakly supervised data. 

The performance of our model on the more challenging CWQ dataset is still promising, where we achieve best performance among the SR-based baselines, except against ToG\textsubscript{GPT-4}. This indicates that even though more complex questions would require better subgraph retrieval ability, the GSR model is still competitive enough to go head-to-head against much larger LLMs. However, the performance gap between all SR-based methods, including ours, against SOTA SP-based methods is still large, indicating that SP-based methods are still dominating complex KGQA. This can be attributed to the fact that SR-based methods, including the SOTA RoG, still struggle with following some specific constraints like `less than' limiting their overall performance.\footnote{More details can be found in Appendix~\ref{app:limitation_constraint}.}

\section{Analysis}
We analyse in detail the effectiveness of several variants of our proposed method, to answer the following research questions. 
\textbf{RQ1}: How to efficiently and effectively prompt LLMs with a retrieved subgraph? (Sec~\ref{sec:rq1})
\textbf{RQ2}: How does the indexing data contribute to the performance on subgraph retrieval?
\textbf{RQ3}: What is the performance upper-bound of the proposed methods? (Sec~\ref{sec:rq4})

\begin{table}[t]
    \centering
    \small
    \begin{tabular}{c|cccc}
        \toprule
        \textbf{Dataset} & \textbf{Hits@1} & \textbf{Hits} & \textbf{F1} & \textbf{Avg. Tokens} \\ \midrule
        \multicolumn{5}{c}{\textit{Subgraph as Reasoning Paths}} \\ \midrule
        WebQSP & 87.8 & 89.6 & 80.1 & 784.5\\
        CWQ & 67.5 & 71.1 & 64.4 & 910.4\\ \midrule
        \multicolumn{5}{c}{\textit{Subgraph as Triple Sets}} \\ \midrule
        WebQSP & 87.3 & 89.7 & 79.8 & 699.3 \\
        CWQ & 66.5 & 70.2 & 63.5 & 777.5\\
        \bottomrule
    \end{tabular}
    \caption{KGQA performance with different ways of prompting LLM with retrieved subgraph.}
    \label{tab:prompt_way}
\end{table}
\begin{figure}[t]
  \centering
  \begin{subfigure}[b]{0.9\linewidth}
        \centering
        \includegraphics[width=\textwidth]{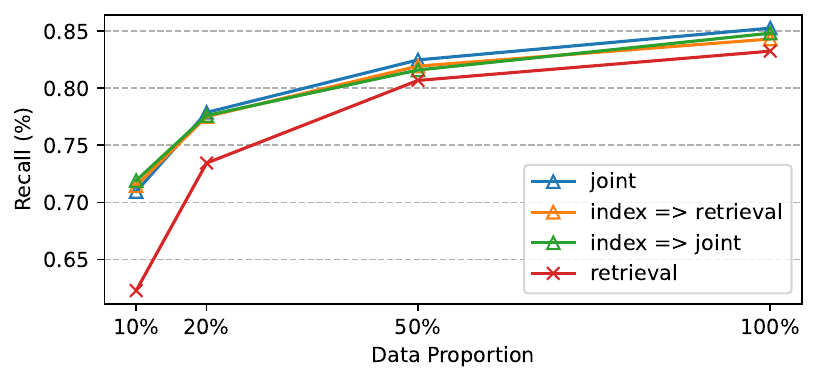}
        \caption{WebQSP}
  \end{subfigure}
  \hfill
  \begin{subfigure}[b]{0.9\linewidth}
        \centering
        \includegraphics[width=\textwidth]{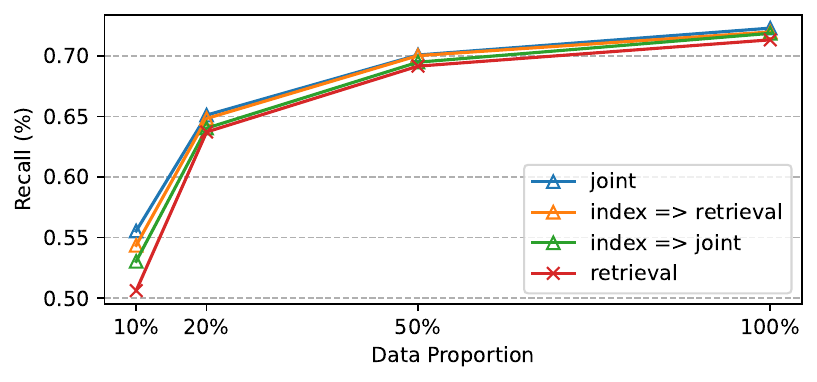}
        \caption{CWQ}
  \end{subfigure}
  \caption{Subgraph retrieval result under low resource KGQA setting. We set beam size $k$=3 here.}
  \label{fig:training_strategy}
\end{figure}
\subsection{How to represent retrieved subgraph}
\label{sec:rq1}
We conduct an in-depth analysis to identify the most suitable format for representing a retrieved subgraph for the LLM reader. Table~\ref{tab:prompt_way} shows how different ways of prompting an LLM reader
can affect the final performance. We observe that representing a retrieved subgraph as reasoning paths generally works better than representing it as triples. Prompting LLMs with triples is more compact.\footnote{Appendix~\ref{app:case_studies} shows an example of why triples are more compact than paths.} As such it offers higher chances to cover an answer if the LLM's context window is limited. This is also demonstrated by the competitive performance on WebQSP. However, the compressed triples' form can easily result in relevant triples, that belong to the same reasoning path, ending up far away in the context of this serialised representation, making the reasoning task harder for the LLM. The effect of this drawback is more evident on the more complex CWQ dataset, where the performance gap between the two approaches becomes larger. Nonetheless, depending on the use-case, using triples as input format can navigate a healthy trade-off between performance and efficiency.

\subsection{The effectiveness of Indexing Data}
\label{sec:rq2}
We evaluate the effectiveness of the indexing step. In addition, we explore different training strategies along the joint-training strategy (discussed in Section~\ref{sec:train}) for the GSR model using the indexing and retrieval data. 1) We directly finetune GSR with retrieval data without indexing data (\textit{retrieval}); 2) we first pretrain GSR with indexing data, and then finetune it with retrieval data (\textit{index => retrieval}). 3) we first pretrain GSR with indexing data, and then finetune it jointly with indexing and retrieval data (\textit{index => joint}). To better highlight the importance of the indexing data, we further define low resource settings, by limiting the availability of retrieval data to 10\%, 20\%, 50\% and 100\%. We use GSR-base in this part for these experiments.

Figure~\ref{fig:training_strategy} shows the recall of GSR-base under different training settings according to the proportion of retrieval data that is trained on. Generally, training the GSR model with retrieval data brings in competitive performance when 100\% of the data is used, but has a large gap with variants using the indexing data when the retrieval data is limited---the gap increases as the data proportion decreases. With only the indexing step, the trained GSR model is able to perform the simplest subgraph retrieval task where the maximum relation hop is 1, with a recall of 42.54 on WebQSP and 21.18 on CWQ. Joint model training with indexing and retrieval data contributes to the best performance, especially on the harder CWQ dataset. This observation is in line with findings from \citet{allenzhu2023physics}, that including finetuning-relevant data instances during pre-training can benefit the model to make better use of the learned relation-level knowledge.

\subsection{Evaluation Results on Full Freebase}
\label{sec:rq4}
To further investigate the upper bound of the proposed GSR method, we apply a full Freebase setting to fetch subgraph based on the subgraph ID from the full Freebase instead of the simplified subgraph (Freebase-SG). Table~\ref{tab:full_res} shows the evaluation results. It is intuitive that the recall increases on both datasets when fetching subgraphs from the full Freebase. On the other hand, the precision on CWQ also increases, which can be attributed to the low question coverage of Freebase-SG that at least 20\% of the questions will always have precision 0 even if the subgraph ID is correct. 
\begin{table}[]
    \centering    \small
    \begin{tabular}{l|p{0.5cm}p{0.5cm}p{0.6cm}|p{0.5cm}p{0.5cm}p{0.5cm}}
    \toprule
        \multirow{2}{*}{\bf{KG}} & \multicolumn{3}{c}{\bf{WebQSP}} & \multicolumn{3}{|c}{\bf{CWQ}} \\ 
        & P & R & F1 & P & R & F1 \\ \midrule
        Freebase-SG & 54.5 & 85.6 & 59.0 & 23.0 & 73.0 & 27.6 \\
        Freebase & 50.2 & 87.5 & 55.2 & 24.9 & 76.6 & 29.7 \\ %
        \bottomrule
    \end{tabular}
    \caption{Subgraph retrieval performance under different KGs.}
    \label{tab:full_res}
\end{table}

\section{Conclusion}
We propose a retrieval-augmented generation framework for KGQA comprising a subgraph retriever and an LLM reader.
 To efficiently and effectively retrieve a subgraph from KG, we propose a novel generative subgraph retrieval method that transform the subgraph retrieval task into a sequential subgraph ID generation task. Our base GSR model with 220M parameters is capable of outperforming the previous 7B SOTA baseline in subgraph retrieval, with around 30$\times$ less parameters. Combined with our LLM reader, we achieve new SOTA performance in subgraph retrieval-based KGQA on both the WebQSP and CWQ datasets. In summary, our proposed GSR model with \textbf{\emph{less}} model parameters, \textbf{\emph{less}} training samples, and learned to generate \textbf{\emph{less}} tokens, achieves \textbf{\emph{more}} efficiency and \textbf{\emph{more}} effectiveness compared SOTA SR-based KGQA works.

\section*{Acknowledgement}
This work is supported by UKRI (grant number EP/S022481/1),  The University of Edinburgh and  Huawei’s Dean’s Funding (C-00006589).

\section*{Limitations}
While our proposed approach achieves SOTA performance among SR-based KGQA methods, there are still limitations to our works. 

First, SR-based KGQA approaches trained using weakly supervised retrieval data still struggle with following special constraints in the questions, since the shortest path between topic entity and answer entity usually does not contain context or entity-type restrictions. Additionally, SR-based KGQA cannot handle questions where the topic entity does not appear at the start of the reasoning chain. These limitations can affect the overall performance of the KGQA task, particularly when the question is complex. In our work, we do not attempt to fix this common limitation in SR-based KGQA approaches, and we consider this as future work.

Second, in this work, we assume the topic entity is already known, which is not typically the case in real-world KGQA tasks. Given that we decompose the subgraph ID as entity and relation chain, there could be accumulative error in cases where we can not identify the correct topic entity.For real-world applications, our proposed method is designed to seamlessly integrate with existing off-the-shelf topic entity linking tools to identify topic entities in the input natural language question, thereby mitigating this limitation.

\bibliography{custom, anthology}

\begin{thebibliography}{35}
\providecommand{\natexlab}[1]{#1}

\bibitem[{Allen-Zhu and Li(2024)}]{allenzhu2023physics}
Zeyuan Allen-Zhu and Yuanzhi Li. 2024.
\newblock \href {https://openreview.net/forum?id=5x788rqbcj} {Physics of language models: Part 3.1, knowledge storage and extraction}.
\newblock In \emph{Forty-first International Conference on Machine Learning}.

\bibitem[{Baek et~al.(2023)Baek, Aji, and Saffari}]{baek-etal-2023-knowledge}
Jinheon Baek, Alham~Fikri Aji, and Amir Saffari. 2023.
\newblock \href {https://doi.org/10.18653/v1/2023.nlrse-1.7} {Knowledge-augmented language model prompting for zero-shot knowledge graph question answering}.
\newblock In \emph{Proceedings of the 1st Workshop on Natural Language Reasoning and Structured Explanations (NLRSE)}, pages 78--106, Toronto, Canada. Association for Computational Linguistics.

\bibitem[{Chen et~al.(2023)Chen, Liu, He, Sun, and Sun}]{chen-etal-2023-understanding}
Xiaoyang Chen, Yanjiang Liu, Ben He, Le~Sun, and Yingfei Sun. 2023.
\newblock \href {https://doi.org/10.18653/v1/2023.findings-acl.681} {Understanding differential search index for text retrieval}.
\newblock In \emph{Findings of the Association for Computational Linguistics: ACL 2023}, pages 10701--10717, Toronto, Canada. Association for Computational Linguistics.

\bibitem[{Das et~al.(2021)Das, Zaheer, Thai, Godbole, Perez, Lee, Tan, Polymenakos, and McCallum}]{das-etal-2021-case}
Rajarshi Das, Manzil Zaheer, Dung Thai, Ameya Godbole, Ethan Perez, Jay~Yoon Lee, Lizhen Tan, Lazaros Polymenakos, and Andrew McCallum. 2021.
\newblock \href {https://doi.org/10.18653/v1/2021.emnlp-main.755} {Case-based reasoning for natural language queries over knowledge bases}.
\newblock In \emph{Proceedings of the 2021 Conference on Empirical Methods in Natural Language Processing}, pages 9594--9611, Online and Punta Cana, Dominican Republic. Association for Computational Linguistics.

\bibitem[{Dettmers et~al.(2023)Dettmers, Pagnoni, Holtzman, and Zettlemoyer}]{dettmers2023qlora}
Tim Dettmers, Artidoro Pagnoni, Ari Holtzman, and Luke Zettlemoyer. 2023.
\newblock \href {https://openreview.net/forum?id=OUIFPHEgJU} {{QL}o{RA}: Efficient finetuning of quantized {LLM}s}.
\newblock In \emph{Thirty-seventh Conference on Neural Information Processing Systems}.

\bibitem[{Guti{\'{e}}rrez et~al.(2024)Guti{\'{e}}rrez, Shu, Gu, Yasunaga, and Su}]{GSGYS2024}
Bernal~Jim{\'{e}}nez Guti{\'{e}}rrez, Yiheng Shu, Yu~Gu, Michihiro Yasunaga, and Yu~Su. 2024.
\newblock \href {https://doi.org/10.48550/ARXIV.2405.14831} {Hipporag: Neurobiologically inspired long-term memory for large language models}.
\newblock \emph{CoRR}, abs/2405.14831.

\bibitem[{He et~al.(2021)He, Lan, Jiang, Zhao, and Wen}]{DBLP:conf/wsdm/HeL0ZW21}
Gaole He, Yunshi Lan, Jing Jiang, Wayne~Xin Zhao, and Ji{-}Rong Wen. 2021.
\newblock \href {https://doi.org/10.1145/3437963.3441753} {Improving multi-hop knowledge base question answering by learning intermediate supervision signals}.
\newblock In \emph{{WSDM} '21, The Fourteenth {ACM} International Conference on Web Search and Data Mining, Virtual Event, Israel, March 8-12, 2021}, pages 553--561. {ACM}.

\bibitem[{Hu et~al.(2022)Hu, Wu, Shu, and Qu}]{hu-etal-2022-logical}
Xixin Hu, Xuan Wu, Yiheng Shu, and Yuzhong Qu. 2022.
\newblock \href {https://aclanthology.org/2022.coling-1.145} {Logical form generation via multi-task learning for complex question answering over knowledge bases}.
\newblock In \emph{Proceedings of the 29th International Conference on Computational Linguistics}, pages 1687--1696, Gyeongju, Republic of Korea. International Committee on Computational Linguistics.

\bibitem[{Huang et~al.(2023)Huang, Yu, Ma, Zhong, Feng, Wang, Chen, Peng, Feng, Qin, and Liu}]{huang2023survey}
Lei Huang, Weijiang Yu, Weitao Ma, Weihong Zhong, Zhangyin Feng, Haotian Wang, Qianglong Chen, Weihua Peng, Xiaocheng Feng, Bing Qin, and Ting Liu. 2023.
\newblock \href {https://doi.org/10.48550/ARXIV.2311.05232} {A survey on hallucination in large language models: Principles, taxonomy, challenges, and open questions}.
\newblock \emph{CoRR}, abs/2311.05232.

\bibitem[{Huang et~al.(2024)Huang, Zhou, Lapata, Vougiouklis, Montella, and Pan}]{huang2024prompting}
Wenyu Huang, Guancheng Zhou, Mirella Lapata, Pavlos Vougiouklis, S{\'{e}}bastien Montella, and Jeff~Z. Pan. 2024.
\newblock \href {https://doi.org/10.48550/ARXIV.2405.06524} {Prompting large language models with knowledge graphs for question answering involving long-tail facts}.
\newblock \emph{CoRR}, abs/2405.06524.

\bibitem[{Jiang et~al.(2023)Jiang, Zhou, Zhao, and Wen}]{jiang2023unikgqa}
Jinhao Jiang, Kun Zhou, Xin Zhao, and Ji-Rong Wen. 2023.
\newblock \href {https://openreview.net/forum?id=Z63RvyAZ2Vh} {Uni{KGQA}: Unified retrieval and reasoning for solving multi-hop question answering over knowledge graph}.
\newblock In \emph{The Eleventh International Conference on Learning Representations}.

\bibitem[{Luo et~al.(2024{\natexlab{a}})Luo, E, Tang, Peng, Guo, Zhang, Ma, Dong, Song, Lin, Zhu, and Luu}]{luo2023chatkbqa}
Haoran Luo, Haihong E, Zichen Tang, Shiyao Peng, Yikai Guo, Wentai Zhang, Chenghao Ma, Guanting Dong, Meina Song, Wei Lin, Yifan Zhu, and Anh~Tuan Luu. 2024{\natexlab{a}}.
\newblock \href {https://doi.org/10.18653/v1/2024.findings-acl.122} {{C}hat{KBQA}: A generate-then-retrieve framework for knowledge base question answering with fine-tuned large language models}.
\newblock In \emph{Findings of the Association for Computational Linguistics ACL 2024}, pages 2039--2056, Bangkok, Thailand and virtual meeting. Association for Computational Linguistics.

\bibitem[{Luo et~al.(2024{\natexlab{b}})Luo, Li, Haf, and Pan}]{luo2024reasoning}
Linhao Luo, Yuan-Fang Li, Reza Haf, and Shirui Pan. 2024{\natexlab{b}}.
\newblock \href {https://openreview.net/forum?id=ZGNWW7xZ6Q} {Reasoning on graphs: Faithful and interpretable large language model reasoning}.
\newblock In \emph{The Twelfth International Conference on Learning Representations}.

\bibitem[{OpenAI(2023)}]{openai2024gpt4}
OpenAI. 2023.
\newblock \href {https://doi.org/10.48550/ARXIV.2303.08774} {{GPT-4} technical report}.
\newblock \emph{CoRR}, abs/2303.08774.

\bibitem[{Pan et~al.(2023)Pan, Razniewski, Kalo, Singhania, Chen, Dietze, Jabeen, Omeliyanenko, Zhang, Lissandrini, Biswas, de~Melo, Bonifati, Vakaj, Dragoni, and Graux}]{PRKSC2023}
Jeff~Z. Pan, Simon Razniewski, Jan{-}Christoph Kalo, Sneha Singhania, Jiaoyan Chen, Stefan Dietze, Hajira Jabeen, Janna Omeliyanenko, Wen Zhang, Matteo Lissandrini, Russa Biswas, Gerard de~Melo, Angela Bonifati, Edlira Vakaj, Mauro Dragoni, and Damien Graux. 2023.
\newblock \href {https://doi.org/10.4230/TGDK.1.1.2} {Large language models and knowledge graphs: Opportunities and challenges}.
\newblock \emph{{TGDK}}, 1(1):2:1--2:38.

\bibitem[{Pan et~al.(2017{\natexlab{a}})Pan, Vetere, G{\'{o}}mez{-}P{\'{e}}rez, and Wu}]{PVGW2017}
Jeff~Z. Pan, Guido Vetere, Jos{\'{e}}~Manu{\'{e}}l G{\'{o}}mez{-}P{\'{e}}rez, and Honghan Wu, editors. 2017{\natexlab{a}}.
\newblock \href {https://doi.org/10.1007/978-3-319-45654-6} {\emph{Exploiting Linked Data and Knowledge Graphs in Large Organisations}}.
\newblock Springer.

\bibitem[{Pan et~al.(2017{\natexlab{b}})Pan, Calvanese, Eiter, Horrocks, Kifer, Lin, and Zhao}]{PCEH+2017}
J.Z. Pan, D.~Calvanese, T.~Eiter, I.~Horrocks, M.~Kifer, F.~Lin, and Y.~Zhao, editors. 2017{\natexlab{b}}.
\newblock \href {https://doi.org/10.1007/978-3-319-49493-7} {\emph{{Reasoning Web: Logical Foundation of Knowledge Graph Construction and Querying Answering}}}.
\newblock Springer.

\bibitem[{Pan et~al.(2024)Pan, Luo, Wang, Chen, Wang, and Wu}]{10387715}
Shirui Pan, Linhao Luo, Yufei Wang, Chen Chen, Jiapu Wang, and Xindong Wu. 2024.
\newblock \href {https://doi.org/10.1109/TKDE.2024.3352100} {Unifying large language models and knowledge graphs: {A} roadmap}.
\newblock \emph{{IEEE} Trans. Knowl. Data Eng.}, 36(7):3580--3599.

\bibitem[{Pradeep et~al.(2023)Pradeep, Hui, Gupta, Lelkes, Zhuang, Lin, Metzler, and Tran}]{DBLP:conf/emnlp/Pradeep0GLZLM023}
Ronak Pradeep, Kai Hui, Jai Gupta, {\'{A}}d{\'{a}}m~D. Lelkes, Honglei Zhuang, Jimmy Lin, Donald Metzler, and Vinh~Q. Tran. 2023.
\newblock \href {https://doi.org/10.18653/V1/2023.EMNLP-MAIN.83} {How does generative retrieval scale to millions of passages?}
\newblock In \emph{Proceedings of the 2023 Conference on Empirical Methods in Natural Language Processing, {EMNLP} 2023, Singapore, December 6-10, 2023}, pages 1305--1321. Association for Computational Linguistics.

\bibitem[{Shuster et~al.(2021)Shuster, Poff, Chen, Kiela, and Weston}]{shuster-etal-2021-retrieval-augmentation}
Kurt Shuster, Spencer Poff, Moya Chen, Douwe Kiela, and Jason Weston. 2021.
\newblock \href {https://doi.org/10.18653/v1/2021.findings-emnlp.320} {Retrieval augmentation reduces hallucination in conversation}.
\newblock In \emph{Findings of the Association for Computational Linguistics: EMNLP 2021}, pages 3784--3803, Punta Cana, Dominican Republic. Association for Computational Linguistics.

\bibitem[{Sun et~al.(2024)Sun, Xu, Tang, Wang, Lin, Gong, Ni, Shum, and Guo}]{sun2024thinkongraph}
Jiashuo Sun, Chengjin Xu, Lumingyuan Tang, Saizhuo Wang, Chen Lin, Yeyun Gong, Lionel Ni, Heung-Yeung Shum, and Jian Guo. 2024.
\newblock \href {https://openreview.net/forum?id=nnVO1PvbTv} {Think-on-graph: Deep and responsible reasoning of large language model on knowledge graph}.
\newblock In \emph{The Twelfth International Conference on Learning Representations}.

\bibitem[{Talmor and Berant(2018)}]{talmor-berant-2018-web}
Alon Talmor and Jonathan Berant. 2018.
\newblock \href {https://doi.org/10.18653/v1/N18-1059} {The web as a knowledge-base for answering complex questions}.
\newblock In \emph{Proceedings of the 2018 Conference of the North {A}merican Chapter of the Association for Computational Linguistics: Human Language Technologies, Volume 1 (Long Papers)}, pages 641--651, New Orleans, Louisiana. Association for Computational Linguistics.

\bibitem[{Tay et~al.(2022)Tay, Tran, Dehghani, Ni, Bahri, Mehta, Qin, Hui, Zhao, Gupta, Schuster, Cohen, and Metzler}]{DBLP:conf/nips/Tay00NBM000GSCM22}
Yi~Tay, Vinh Tran, Mostafa Dehghani, Jianmo Ni, Dara Bahri, Harsh Mehta, Zhen Qin, Kai Hui, Zhe Zhao, Jai~Prakash Gupta, Tal Schuster, William~W. Cohen, and Donald Metzler. 2022.
\newblock \href {http://papers.nips.cc/paper\_files/paper/2022/hash/892840a6123b5ec99ebaab8be1530fba-Abstract-Conference.html} {Transformer memory as a differentiable search index}.
\newblock In \emph{Advances in Neural Information Processing Systems 35: Annual Conference on Neural Information Processing Systems 2022, NeurIPS 2022, New Orleans, LA, USA, November 28 - December 9, 2022}.

\bibitem[{Tonmoy et~al.(2024)Tonmoy, Zaman, Jain, Rani, Rawte, Chadha, and Das}]{tonmoy2024comprehensive}
S.~M. Towhidul~Islam Tonmoy, S.~M.~Mehedi Zaman, Vinija Jain, Anku Rani, Vipula Rawte, Aman Chadha, and Amitava Das. 2024.
\newblock \href {https://doi.org/10.48550/ARXIV.2401.01313} {A comprehensive survey of hallucination mitigation techniques in large language models}.
\newblock \emph{CoRR}, abs/2401.01313.

\bibitem[{Touvron et~al.(2023)Touvron, Martin, Stone, Albert, Almahairi, Babaei, Bashlykov, Batra, Bhargava, Bhosale, Bikel, Blecher, Canton{-}Ferrer, Chen, Cucurull, Esiobu, Fernandes, Fu, Fu, Fuller, Gao, Goswami, Goyal, Hartshorn, Hosseini, Hou, Inan, Kardas, Kerkez, Khabsa, Kloumann, Korenev, Koura, Lachaux, Lavril, Lee, Liskovich, Lu, Mao, Martinet, Mihaylov, Mishra, Molybog, Nie, Poulton, Reizenstein, Rungta, Saladi, Schelten, Silva, Smith, Subramanian, Tan, Tang, Taylor, Williams, Kuan, Xu, Yan, Zarov, Zhang, Fan, Kambadur, Narang, Rodriguez, Stojnic, Edunov, and Scialom}]{touvron2023llama}
Hugo Touvron, Louis Martin, Kevin Stone, Peter Albert, Amjad Almahairi, Yasmine Babaei, Nikolay Bashlykov, Soumya Batra, Prajjwal Bhargava, Shruti Bhosale, Dan Bikel, Lukas Blecher, Cristian Canton{-}Ferrer, Moya Chen, Guillem Cucurull, David Esiobu, Jude Fernandes, Jeremy Fu, Wenyin Fu, Brian Fuller, Cynthia Gao, Vedanuj Goswami, Naman Goyal, Anthony Hartshorn, Saghar Hosseini, Rui Hou, Hakan Inan, Marcin Kardas, Viktor Kerkez, Madian Khabsa, Isabel Kloumann, Artem Korenev, Punit~Singh Koura, Marie{-}Anne Lachaux, Thibaut Lavril, Jenya Lee, Diana Liskovich, Yinghai Lu, Yuning Mao, Xavier Martinet, Todor Mihaylov, Pushkar Mishra, Igor Molybog, Yixin Nie, Andrew Poulton, Jeremy Reizenstein, Rashi Rungta, Kalyan Saladi, Alan Schelten, Ruan Silva, Eric~Michael Smith, Ranjan Subramanian, Xiaoqing~Ellen Tan, Binh Tang, Ross Taylor, Adina Williams, Jian~Xiang Kuan, Puxin Xu, Zheng Yan, Iliyan Zarov, Yuchen Zhang, Angela Fan, Melanie Kambadur, Sharan Narang, Aur{\'{e}}lien Rodriguez, Robert Stojnic, Sergey Edunov,
  and Thomas Scialom. 2023.
\newblock \href {https://doi.org/10.48550/ARXIV.2307.09288} {Llama 2: Open foundation and fine-tuned chat models}.
\newblock \emph{CoRR}, abs/2307.09288.

\bibitem[{Wang et~al.(2024{\natexlab{a}})Wang, Chen, Hu, Yang, Liu, Shen, Wei, Zhang, Gu, Zhou, Pan, Zhang, and Chen}]{WCHY+2024}
Junjie Wang, Mingyang Chen, Binbin Hu, Dan Yang, Ziqi Liu, Yue Shen, Peng Wei, Zhiqiang Zhang, Jinjie Gu, Jun Zhou, Jeff~Z Pan, Wen Zhang, and Huajun Chen. 2024{\natexlab{a}}.
\newblock {Learning to Plan for Retrieval-Augmented Large Language Models from Knowledge Graphs}.
\newblock In \emph{Proc. of Empirical Methods in Natural Language Processing}.

\bibitem[{Wang et~al.(2024{\natexlab{b}})Wang, Lipka, Rossi, Siu, Zhang, and Derr}]{DBLP:conf/aaai/0160LRSZD24}
Yu~Wang, Nedim Lipka, Ryan~A. Rossi, Alexa~F. Siu, Ruiyi Zhang, and Tyler Derr. 2024{\natexlab{b}}.
\newblock \href {https://doi.org/10.1609/AAAI.V38I17.29889} {Knowledge graph prompting for multi-document question answering}.
\newblock In \emph{Thirty-Eighth {AAAI} Conference on Artificial Intelligence, {AAAI} 2024, Thirty-Sixth Conference on Innovative Applications of Artificial Intelligence, {IAAI} 2024, Fourteenth Symposium on Educational Advances in Artificial Intelligence, {EAAI} 2014, February 20-27, 2024, Vancouver, Canada}, pages 19206--19214. {AAAI} Press.

\bibitem[{Wu et~al.(2024)Wu, Huang, Hu, Hua, Qi, Chen, and Pan}]{WHHH+2024}
Yike Wu, Yi~Huang, Nan Hu, Yuncheng Hua, Guilin Qi, Jiaoyan Chen, and Jeff~Z. Pan. 2024.
\newblock {CoTKR: Chain-of-Thought Enhanced Knowledge Rewriting for Complex Knowledge Graph Question Answering}.
\newblock In \emph{Proc. of Empirical Methods in Natural Language Processing}.

\bibitem[{Yih et~al.(2016)Yih, Richardson, Meek, Chang, and Suh}]{yih-etal-2016-value}
Wen-tau Yih, Matthew Richardson, Chris Meek, Ming-Wei Chang, and Jina Suh. 2016.
\newblock \href {https://doi.org/10.18653/v1/P16-2033} {The value of semantic parse labeling for knowledge base question answering}.
\newblock In \emph{Proceedings of the 54th Annual Meeting of the Association for Computational Linguistics (Volume 2: Short Papers)}, pages 201--206, Berlin, Germany. Association for Computational Linguistics.

\bibitem[{Yu et~al.(2023)Yu, Zhang, Ng, Zhu, Li, Wang, Hu, Wang, Wang, and Xiang}]{yu2023decaf}
Donghan Yu, Sheng Zhang, Patrick Ng, Henghui Zhu, Alexander~Hanbo Li, Jun Wang, Yiqun Hu, William~Yang Wang, Zhiguo Wang, and Bing Xiang. 2023.
\newblock \href {https://openreview.net/forum?id=XHc5zRPxqV9} {Dec{AF}: Joint decoding of answers and logical forms for question answering over knowledge bases}.
\newblock In \emph{The Eleventh International Conference on Learning Representations}.

\bibitem[{Zhang et~al.(2022)Zhang, Zhang, Yu, Tang, Tang, Li, and Chen}]{zhang-etal-2022-subgraph}
Jing Zhang, Xiaokang Zhang, Jifan Yu, Jian Tang, Jie Tang, Cuiping Li, and Hong Chen. 2022.
\newblock \href {https://doi.org/10.18653/v1/2022.acl-long.396} {Subgraph retrieval enhanced model for multi-hop knowledge base question answering}.
\newblock In \emph{Proceedings of the 60th Annual Meeting of the Association for Computational Linguistics (Volume 1: Long Papers)}, pages 5773--5784, Dublin, Ireland. Association for Computational Linguistics.

\bibitem[{Zhang et~al.(2023{\natexlab{a}})Zhang, Zhang, Ke, Li, Huang, Shao, Cao, and Lv}]{DBLP:journals/aiopen/ZhangZKLHSCL23}
Lingxi Zhang, Jing Zhang, Xirui Ke, Haoyang Li, Xinmei Huang, Zhonghui Shao, Shulin Cao, and Xin Lv. 2023{\natexlab{a}}.
\newblock \href {https://doi.org/10.1016/J.AIOPEN.2022.12.003} {A survey on complex factual question answering}.
\newblock \emph{{AI} Open}, 4:1--12.

\bibitem[{Zhang et~al.(2023{\natexlab{b}})Zhang, Li, Cui, Cai, Liu, Fu, Huang, Zhao, Zhang, Chen, Wang, Luu, Bi, Shi, and Shi}]{zhang2023sirens}
Yue Zhang, Yafu Li, Leyang Cui, Deng Cai, Lemao Liu, Tingchen Fu, Xinting Huang, Enbo Zhao, Yu~Zhang, Yulong Chen, Longyue Wang, Anh~Tuan Luu, Wei Bi, Freda Shi, and Shuming Shi. 2023{\natexlab{b}}.
\newblock \href {https://doi.org/10.48550/ARXIV.2309.01219} {Siren's song in the {AI} ocean: {A} survey on hallucination in large language models}.
\newblock \emph{CoRR}, abs/2309.01219.

\bibitem[{Zhang et~al.(2018)Zhang, Dai, Kozareva, Smola, and Song}]{zhang2017variational}
Yuyu Zhang, Hanjun Dai, Zornitsa Kozareva, Alexander~J. Smola, and Le~Song. 2018.
\newblock \href {https://doi.org/10.1609/AAAI.V32I1.12057} {Variational reasoning for question answering with knowledge graph}.
\newblock In \emph{Proceedings of the Thirty-Second {AAAI} Conference on Artificial Intelligence, (AAAI-18), the 30th innovative Applications of Artificial Intelligence (IAAI-18), and the 8th {AAAI} Symposium on Educational Advances in Artificial Intelligence (EAAI-18), New Orleans, Louisiana, USA, February 2-7, 2018}, pages 6069--6076. {AAAI} Press.

\bibitem[{Zhuang et~al.(2022)Zhuang, Ren, Shou, Pei, Gong, Zuccon, and Jiang}]{DBLP:journals/corr/abs-2206-10128}
Shengyao Zhuang, Houxing Ren, Linjun Shou, Jian Pei, Ming Gong, Guido Zuccon, and Daxin Jiang. 2022.
\newblock \href {https://doi.org/10.48550/ARXIV.2206.10128} {Bridging the gap between indexing and retrieval for differentiable search index with query generation}.
\newblock \emph{CoRR}, abs/2206.10128.

\end{thebibliography}

\appendix

\section{Implementation Details}
\label{app:implementation}
\subsection{Environment}
We conduct all the experiments on a single NVIDIA A100 80G GPU. For all GPT-4 API calls, we use \texttt{gpt-4-turbo-preview} as the API endpoint. 
\subsection{Training LLM Reader}
For training LLaMA2 and LLaMA3 reader, we use the trained GSR model to generate subgraph ID for all questions in WebQSP and CWQ training set, with beam size $k$=10 and fetching subgraph from simplified Freebase with top-3 valid relation chain. Then we use prompt provided in Appendix~\ref{app:prompt} to generate supervised finetune data for training LLM readers.
\subsection{Hyperparameters}
For all GSR models (based on T5\footnote{\url{https://huggingface.co/google-t5}}), we set the epoch numbers to 50. We set the batch size to \{128, 128, 64\} and learning rate to \{5e-4, 2e-4, 1e-4\} for GSR-base, GSR-large and GSR-3b respectively. For LLM readers, we set epoch numbers to 3, learning rate to 2e-4, and max sequence length to 4,096 for both LLaMA2\footnote{\url{https://huggingface.co/unsloth/llama-2-7b-chat-bnb-4bit}}and LLaMA3\footnote{\url{https://huggingface.co/unsloth/llama-3-8b-Instruct-bnb-4bit}}. The QLora hyperparameters is set to $r=16$ and $\alpha=16$, with 4 bit quantization.
\subsection{Experiment Cost}
\paragraph{API Cost}
We utilise GPT-4 API in both indexing data creation and retrieval data creation. For generating pseudo questions as indexing data, we spent around \$40 US dollars. While for distil knowledge from GPT-4 to filter retrieval data, we spent around \$35 US dollars. 
\paragraph{Training Cost}
Table~\ref{tab:training_cost} shows the training cost of both GSR retriever and LLM reader used in this work.
\begin{table}[t]
    \centering
    \begin{tabular}{l|c}
    \toprule
        \textbf{Model} & \textbf{Training Time} \\ \midrule
        \multicolumn{2}{c}{\textit{Subgraph Retriever}} \\ \midrule
        GSR-base & 2 hours \\ 
        GSR-large & 8 hours \\ 
        GSR-3B & 20 hours \\ \midrule
        RoG + T5-base & 2 hours \\
        \midrule
        \multicolumn{2}{c}{\textit{LLM Reader}} \\ \midrule
        LLaMA2-7B & 10 hours\\
        LLaMA3-8B & 12 hours\\
        \bottomrule
    \end{tabular}
    \caption{Training time spent on single NVIDIA A100 80G GPU. GSR models are jointly trained with indexing data and GPT-selected retreival data. RoG+T5-base is trained with GPT-selected retrieval data only.}
    \label{tab:training_cost}
\end{table}
\paragraph{Inference Cost}
Table~\ref{tab:inference_cost} shows the inference cost of subgraph retrieval step on both datasets. 
\begin{table}[t]
    \centering
    \small
    \begin{tabular}{l|c|cc}
    \toprule
        Models & Params & WebQSP & CWQ \\ \midrule
        RoG & 7B & 1,650s & 3,823s\\ \midrule
        RoG+FlanT5-base & 0.2B & 912s & 1,887s \\ \midrule
        GSR-base & 0.2B & 117s & 258s \\
        GSR-large & 0.8B & 190s & 443s \\
        GSR-3B & 3B & 217s & 490s \\
    \bottomrule
    \end{tabular}
    \caption{Inference cost on both datasets. Measured on single NVIDIA A100 80G GPU.}
    \label{tab:inference_cost}
\end{table}

\section{Dataset details}
\label{app:datasets}
\subsection{Original KGQA Dataset}
In this work, we use the WebQSP and CWQ datasets processed by \citet{luo2024reasoning} for all the experiments. Following previous studies \citep{zhang-etal-2022-subgraph, luo2024reasoning}, we utilize the subgraph of Freebase (Freebase-SG) instead of the full Freebase for most of the experiments to enhance efficiency. Table~\ref{tab:data_statistic_ori} shows the statistic of the original dataset. Coverage refers to the question coverage rate of Freebase-SG, which is the proportion of questions where at least one answer entity appears in Freebase-SG.

\begin{table}[]
    \centering
    \small
    \begin{tabular}{c|cccc}
        \toprule
        \textbf{Dataset} & \textbf{Train} & \textbf{Test} & \textbf{Max Hop} & \textbf{Coverage} \\ \midrule
        WebQSP & 2,826 & 1,628 & 2 & 94.9\% \\
        CWQ & 27,639 & 3,531 & 4 & 79.3\% \\ \bottomrule
    \end{tabular}
    \caption{Statistic of original dataset.}
    \label{tab:data_statistic_ori}
\end{table}

\subsection{Training Data}
\begin{figure}
    \centering
    \includegraphics[width=0.46\textwidth]{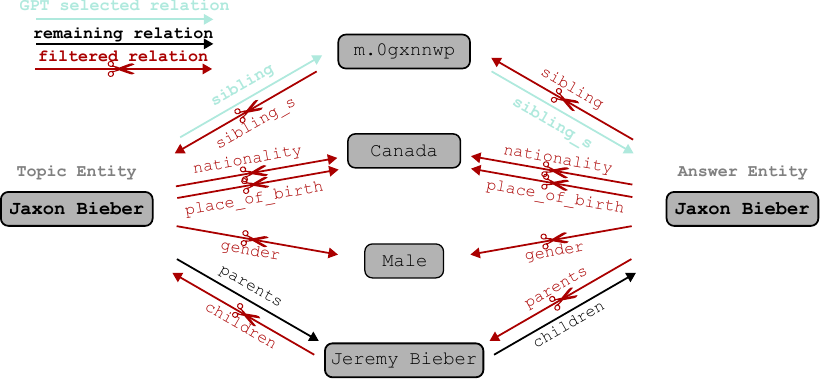}
    \caption{An example of filtering weakly supervised data. The example is taken from WebQSP training set with question: what is the name of justin bieber brother?}
    \label{fig:data_creation}
\end{figure}

\paragraph{Indexing Data}
For each relation, we prompt GPT-4 to generate 10 pseudo questions with placeholder \texttt{[SUBJECT]}. In some cases, the generation is invalid, where the model fails to follow our instruction of generating placeholder. At such cases, we simply remove the generated pseudo question. Finally, we created the indexing data with 83,104 pseudo questions to relation ID mapping for a total of 8,321 relations. Noted that we filter out other relations in Freebase since we can not find any valid triples related to these relations, which means these relations will never contribute to the subgraph retrieval task.
\paragraph{Retrieval Data}
Table~\ref{tab:data_statistic_train} shows the statistic of different types of retrieval data. While Figure~\ref{fig:data_creation} shows an example of how we process the weakly supervised data to get the filtered data and GPT-selected data.
\begin{table}[]
    \centering
    \begin{tabular}{c|cc}
        \toprule
        \bf{Data Type} & \bf{WebQSP} & \bf{CWQ} \\ \midrule
        Raw & 9,745 & 87,420 \\
        Filtered & 5,551 & 46,783 \\
        GPT-select & 3,741 & 31,035 \\ \bottomrule
    \end{tabular}
    \caption{Statistic of retrieval data.}
    \label{tab:data_statistic_train}
\end{table}

\section{Prompt Details}
\label{app:prompt}
Prompt to clean the weakly supervised dataset:

\begin{tcolorbox}
    Given the question and candidate relation reasoning path, select the path that is most helpful for reaching the answer of the question. Only return the number list separated with comma of the relation path without anything else.

    Paths:
    \textcolor{red}{\{path\_list\}}
    
    Question:
    \textcolor{red}{\{question\}}
\end{tcolorbox}

\noindent Prompt to generate pseudo questions:

\begin{tcolorbox}
    Given the Freebase relation \textcolor{red}{\{relation\}} and a triple example of the relation \textcolor{red}{\{triple\_example\}}, generate 10 templates that can be used to ask question about the relation. Use [SUBJECT] to identify subject entity (not the one in the example)
\end{tcolorbox}

\noindent Prompt for LLaMA reader (reasoning paths):

\begin{tcolorbox}
    Based on the reasoning paths, please answer the given question. Please keep the answer as simple as possible and return all the possible answers as a list.
    
    Reasoning Paths: \textcolor{red}{\{reasoning\_paths\}}
    
    Question: \textcolor{red}{\{question\}}
\end{tcolorbox}

\noindent Prompt for LLaMA reader (triple sets):

\begin{tcolorbox}
    Based on the KG triples, please answer the given question. Please keep the answer as simple as possible and return all the possible answers as a list.
    
    KG Triples: \textcolor{red}{\{subgraph\_triples\}}
    
    Question: \textcolor{red}{\{question\}}
\end{tcolorbox}

\section{Detailed Results}
\subsection{How many beam size do we need}
\label{app:beam}
\begin{figure}[t]
  \centering
  \begin{subfigure}[b]{0.48\textwidth}
        \centering
        \includegraphics[width=\textwidth]{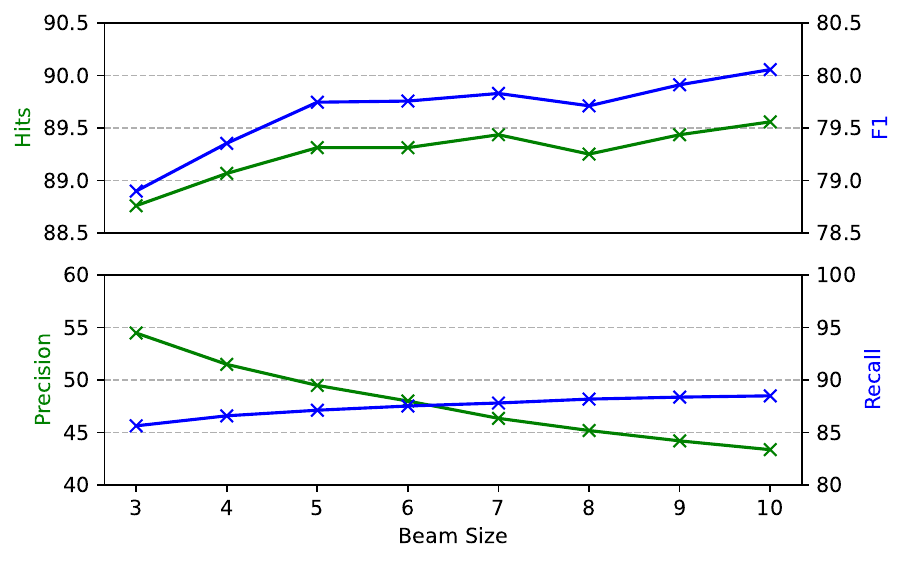}
        \caption{WebQSP}
  \end{subfigure}
  \hfill
  \begin{subfigure}[b]{0.48\textwidth}
        \centering
        \includegraphics[width=\textwidth]{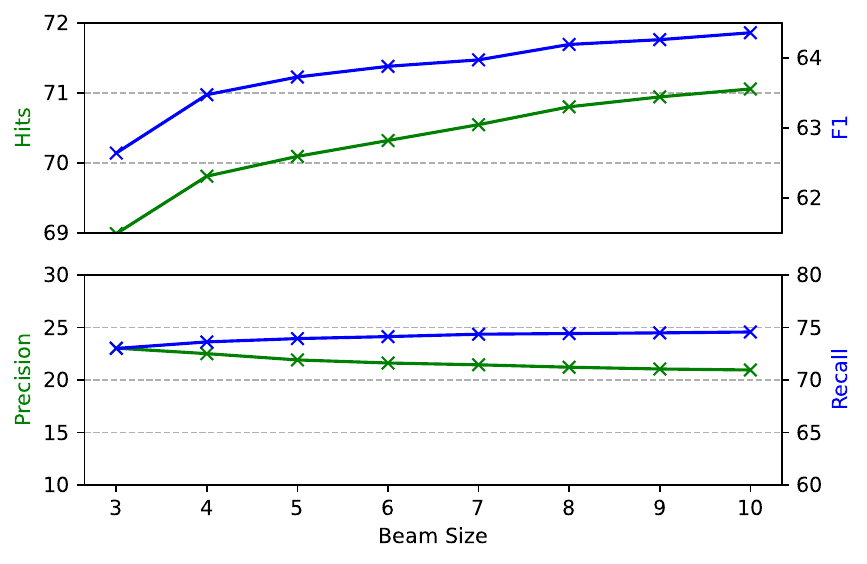}
        \caption{CWQ}
  \end{subfigure}
  \caption{Subgraph retrieval and end-to-end KGQA performance based on different Beam size $k$. Upper figure measures end-to-end metrics (Hits and F1), while bottom figure measures subgraph retrieval metrics (Precision and Recall).}
  \label{fig:beam_size}
\end{figure}
In this part we discuss how the different beam size affect both subgraph retrieval performance and end-to-end KGQA performance. Noted that no matter how large the beam size $k$ is, we always keep the max size of valid subgraph IDs to 3. Figure~\ref{fig:beam_size} shows the evaluation results when changing beam size from 3 to 10. From the result, we can observe a clear precision drop as well as little recall improvement on subgraph retrieval performance on WebQSP dataset. While for CWQ dataset, we can find the drop and improvement are more balanced. The `seesaw' of precision and recall then affect the end-to-end KGQA performance. On WebQSP dataset, when we increase $k$, the performance is quite unstable, since we bring a lot more noise with lower precision. While on the more complex CWQ dataset, the performance improvement along with $k$ is clear and consistent. This findings provide an insight that for the choice of $k$ is conditioned on the complexity of the task, which we should choose a relatively low $k$ for simple task while a larger $k$ for hard task.

\subsection{End-to-end results with different training data}
Table~\ref{tab:end_to_end_res_ext} shows extended end-to-end evaluation results under GSR models trained with different retrieval data.
\begin{table}[t]
    \centering
    \small
    \begin{tabular}{l|p{0.5cm}p{0.5cm}p{0.6cm}|p{0.5cm}p{0.5cm}p{0.5cm}}
    \toprule
        \multirow{2}{*}{\bf{Model}} & \multicolumn{3}{c}{\bf{WebQSP}} & \multicolumn{3}{|c}{\bf{CWQ}} \\ 
        & H@1 & Hits & F1 & H@1 & Hits & F1 \\ \midrule
        \multicolumn{7}{c}{\textit{ours w/ raw data}} \\ \midrule
        GSR-3b & 87.2 & 89.1 & 78.9 & 63.7 & 66.9 & 60.3 \\ \midrule
        \multicolumn{7}{c}{\textit{ours w/ filtered data}} \\ \midrule
        GSR-3b & 86.8 & 89.2 & 79.3 & 64.2 & 67.7 & 61.2\\ \midrule
        \multicolumn{7}{c}{\textit{ours w/ GPT selected data}} \\ \midrule
        GSR-3b & \textbf{87.8} & \textbf{89.6} & \textbf{80.1} & \textbf{67.5} & \textbf{71.1} & \textbf{64.4} \\ 
        \bottomrule
    \end{tabular}
\caption{End-to-end KGQA performance with LLM reader. H@1 stands for Hits@1.}
    \label{tab:end_to_end_res_ext}
\end{table}

\subsection{Subgraph retrieval results under $k$=10}
\label{app:top10}
Table~\ref{tab:subgraph_ret_res_top10} shows the subgraph retrieval results under beam size $k$=10. Noted that we still limit at most 3 valid paths per question.
\begin{table}[t]
    \centering
    \small
    \begin{tabular}{l|p{0.5cm}p{0.5cm}p{0.6cm}|p{0.5cm}p{0.5cm}p{0.5cm}}
    \toprule
        \multirow{2}{*}{\bf{Model}} & \multicolumn{3}{c}{\bf{WebQSP}} & \multicolumn{3}{|c}{\bf{CWQ}} \\ 
        & P & R & F1 & P & R & F1 \\ \midrule
        \multicolumn{7}{c}{\textit{ours w/ raw data}} \\ \midrule
        GSR-base & 35.26 & 86.97 & 41.79 & 16.24 & 73.79 & 20.40 \\
        GSR-large & 34.08 & 87.91 & 40.43 & 16.61 & 74.27 & 20.73 \\
        GSR-3b & 34.54 & 88.05 & 41.09 & 16.02 & 74.84 & 20.10 \\ \midrule 
        \multicolumn{7}{c}{\textit{ours w/ filtered data}} \\ \midrule
        GSR-base & 35.66 & 88.68 & 43.73 & 16.81 & 73.89 & 21.58 \\
        GSR-large & 36.19 & 88.90 & 44.01 & 17.07 & 73.98 & 21.82 \\
        GSR-3b & 36.52 & 88.20 & 44.01 & 17.00 & 74.06 & 21.64 \\ 
        \midrule
        \multicolumn{7}{c}{\textit{ours w/ GPT selected data}} \\ \midrule
        GSR-base & 45.27 & 87.93 & 51.88 & 20.93 & 73.66 & 25.70 \\
        GSR-large & 45.71 & 87.73 & 51.99 & 21.22 & 73.86 & 25.92 \\
        GSR-3b & 43.35 & 88.48 & 49.92 & 20.95 & 74.56 & 25.61  \\
        \bottomrule
    \end{tabular}
    \caption{Subgraph retrieval results under beam size $k$=10.}
    \label{tab:subgraph_ret_res_top10}
\end{table}

\subsection{Patterns of GPT-selected data}
In this part, we analyse the patterns of GPT-selected data that contribute to the performance improvement compared with raw data. By analysing the differences between raw paths and GPT-selected paths, we found an interesting pattern of excluded paths: GPT-selected paths usually contain less repeated relations (e.g., people.person.gender -> people.person.gender). Around 22\% of raw paths and 32\% of paths excluded by GPT have a repeated relation, but less than 4\% of GPT-selected paths contain any repeated relation. This finding can be used for explaining why the data filtering method works, since most repeated paths have both forward and backward directions (e.g., \textit{topic entity} $\rightarrow$ \textit{gender} $\rightarrow$ \textit{Male} $\leftarrow$ \textit{gender} $\leftarrow$ \textit{answer entity}) and they will be filtered in the filtered data setting.

\section{Generalize to other Datasets}
In this section, we investigate whether the proposed method can be generalized to other datasets and other KGs. In this part, we follow RoG \citep{luo2024reasoning} to use MetaQA datasets \citep{zhang2017variational}. Specifically, MetaQA is a KGQA dataset in the movie domain, based on the WikiMovies KG. To better show the effectiveness of our proposed method and for a fair comparison, we follow the same low resource setting from RoG by using only 1,000 randomly sampled data instances for training, without leveraging the gold annotated relation chain. In addition, we use the 3-hop subset of MetaQA, which requires 3 hops from topic entity to target entity. Table~\ref{tab:metaqa} shows the results, we use fine-tuned LLaMA2 as reader for fair comparison. We use Hits and F1 as the metrics to measure the performance of baseline (RoG) and proposed methods.

\begin{table}[]
    \centering
    \begin{tabular}{c|cc}
    \toprule
        \bf{Model} & \bf{Hits} & \bf{F1} \\
        \midrule
        RoG	& 88.98 & 50.68 \\ \midrule
        GSR-base & 90.27 & 55.82 \\
        GSR-large & 91.12 & 57.41 \\
        GSR-3b & \textbf{92.25} & \textbf{63.02} \\
        \bottomrule

    \end{tabular}
    \caption{End-to-end KGQA performance on MetaQA 3-hop test set. All models are trained with 1,000 training samples.}
    \label{tab:metaqa}
\end{table}

From the results, we can find that the superiority of the proposed method still holds on MetaQA 3-hop data, consistently outperforming RoG (7B) with less model parameters in the same experimental setting (low resources and no gold annotation). 

\section{Case Studies}
\label{app:case_studies}
\begin{table*}[h]
    \small
    \centering
    \begin{tabular}{lp{11cm}}
    \toprule
\multicolumn{2}{c}{who was vp for nixon} \\ \midrule
\multirow{3}{*}{GSR (w/ raw data)} & 1:government.us\_president.vice\_president \\
& 2:government.us\_vice\_president.to\_president \\
& 3:people.person.gender -- people.person.gender \\
\midrule
\multirow{3}{*}{GSR (w/ filtered data)} & 1:government.us\_president.vice\_president \\
& 2:government.us\_vice\_president.to\_president \\
& 3:government.government\_office\_category.officeholders -- government.politician.government\_positions\_held \\
\midrule
\multirow{3}{*}{GSR (w/ GPT-selected data)} & 1:government.us\_president.vice\_president \\
& 2:government.political\_appointer.appointees -- \ \ government.government\_position\_held.office\_holder \\
& 3:common.image.appears\_in\_topic\_gallery -- government.us\_president.vice\_president \\
\midrule
\multicolumn{2}{c}{who did armie hammer play in the social network} \\ \midrule
\multirow{3}{*}{GSR (w/ raw data)} & 1:people.person.profession -- people.person.profession \\
& 2:people.person.nationality -- tv.tv\_program.country\_of\_origin \\
& 3:people.person.nationality -- people.person.nationality \\
\midrule
\multirow{3}{*}{GSR (w/ filtered data)} & 1:people.person.profession \\
& 2:film.actor.film -- film.performance.character \\
& 3:film.actor.film -- film.film\_character.portrayed\_in\_films \\
\midrule
\multirow{3}{*}{GSR (w/ GPT-selected data)} & 1:film.actor.film -- film.performance.character \\
& 2:tv.tv\_actor.starring\_roles -- tv.regular\_tv\_appearance.character \\
& 3:film.film.starring -- film.performance.character \\
\midrule
\multicolumn{2}{c}{The artist that created the art series of Water Lilies was inspired by what?} \\ \midrule
\multirow{3}{*}{GSR (w/ raw data)} & 1:visual\_art.art\_series.artist -- influence.influence\_node.influenced\_by \\
& 2:visual\_art.art\_series.artist -- influence.influence\_node.influenced \\
& 3:visual\_art.artwork.artist -- influence.influence\_node.influenced \\
\midrule
\multirow{3}{*}{GSR (w/ filtered data)} & 1:visual\_art.art\_series.artist -- influence.influence\_node.influenced\_by \\
& 2:visual\_art.artwork.artist -- influence.influence\_node.influenced\_by \\
& 3:visual\_art.artwork.artist -- influence.influence\_node.influenced \\
\midrule
\multirow{3}{*}{GSR (w/ GPT-selected data)} & 1:visual\_art.art\_series.artist -- influence.influence\_node.influenced\_by \\
& 2:visual\_art.artwork.artist -- influence.influence\_node.influenced\_by \\
& 3:visual\_art.art\_series.artworks -- influence.influence\_node.influenced\_by \\
\midrule
\multicolumn{2}{c}{What is the official language of the area where the government of Ukraine is?} \\ \midrule
\multirow{3}{*}{GSR (w/ raw data)} & 1:government.governmental\_jurisdiction.government\_bodies -- location.country.official\_language \\
& 2:government.governmental\_jurisdiction.government\_bodies -- language.human\_language.countries\_spoken\_in \\
& 3:government.governmental\_jurisdiction.government\_bodies -- location.country.languages\_spoken \\
\midrule
\multirow{3}{*}{GSR (w/ filtered data)} & 1:government.governmental\_jurisdiction.government\_bodies -- location.country.official\_language \\
& 2:government.governmental\_jurisdiction.government\_bodies -- location.country.languages\_spoken \\
& 3:government.governmental\_jurisdiction.government\_bodies -- language.human\_language.main\_country \\
\midrule
\multirow{3}{*}{GSR (w/ GPT-selected data)} & 1:government.government.government\_for -- location.country.official\_language \\
& 2:government.government.government\_for -- location.country.languages\_spoken \\
& 3:government.governmental\_body.jurisdiction -- location.country.official\_language \\

    \bottomrule
    \end{tabular}
    \caption{Subgraph ID generation cases.}
    \label{tab:subgraphid_case}
\end{table*}

\begin{table*}[h]
    \small
    \centering
    \begin{tabular}{lp{11cm}}
    \toprule
        Question & Lou Seal is the mascot for the team that last won the World Series when? \\ \midrule
        Subgraph ID & sports.mascot.team -- sports.sports\_championship\_event.champion\\
        & sports.mascot.team -- sports.sports\_team.championships\\
        & sports.sports\_team.team\_mascot -- sports.sports\_team.championships\\\midrule
        Retrived Subgraph & Lou Seal $\rightarrow$ sports.mascot.team $\rightarrow$ San Francisco Giants $\leftarrow$ sports.sports\_championship\_event.champion $\leftarrow$ 2014 World Series\\
        as Paths & Lou Seal $\rightarrow$ sports.mascot.team $\rightarrow$ San Francisco Giants $\leftarrow$ sports.sports\_championship\_event.champion $\leftarrow$ 2010 World Series\\
        & Lou Seal $\rightarrow$ sports.mascot.team $\rightarrow$ San Francisco Giants $\leftarrow$ sports.sports\_championship\_event.champion $\leftarrow$ 2012 World Series\\
        & Lou Seal $\rightarrow$ sports.mascot.team $\rightarrow$ San Francisco Giants $\rightarrow$ sports.sports\_team.championships $\rightarrow$ 2014 World Series \\
        & Lou Seal $\rightarrow$ sports.mascot.team $\rightarrow$ San Francisco Giants $\rightarrow$ sports.sports\_team.championships $\rightarrow$ 2010 World Series \\
        & Lou Seal $\rightarrow$ sports.mascot.team $\rightarrow$ San Francisco Giants $\rightarrow$ sports.sports\_team.championships $\rightarrow$ 2012 World Series \\
        & Lou Seal $\leftarrow$ sports.sports\_team.team\_mascot $\leftarrow$ San Francisco Giants $\rightarrow$ sports.sports\_team.championships $\rightarrow$ 2014 World Series \\
        & Lou Seal $\leftarrow$ sports.sports\_team.team\_mascot $\leftarrow$ San Francisco Giants $\rightarrow$ sports.sports\_team.championships $\rightarrow$ 2010 World Series \\
        & Lou Seal $\leftarrow$ sports.sports\_team.team\_mascot $\leftarrow$ San Francisco Giants $\rightarrow$ sports.sports\_team.championships $\rightarrow$ 2012 World Series \\
        \midrule
        Retrived Subgraph & Lou Seal, sports.mascot.team, San Francisco Giants \\
        as Triples & 2014 World Series, sports.sports\_championship\_event.champion, San Francisco Giants \\
        & 2010 World Series, sports.sports\_championship\_event.champion, San Francisco Giants \\
        & 2012 World Series, sports.sports\_championship\_event.champion, San Francisco Giants \\
        & San Francisco Giants, sports.sports\_team.championships, 2014 World Series \\
        & San Francisco Giants, sports.sports\_team.championships, 2010 World Series \\
        & San Francisco Giants, sports.sports\_team.championships, 2012 World Series \\
        & San Francisco Giants, sports.sports\_team.team\_mascot, Lou Seal\\

        \midrule
        Reference Answer(s) & 2014 World Series \\
        \midrule
        Predicted Answers (Paths) & 2014 World Series \\
        Predicted Answers (Triples) & 2014 World Series \\
    \bottomrule
    \end{tabular}
    \caption{One success example sampled from CWQ test set.}
    \label{tab:case_success}
\end{table*}

\begin{table*}[h]
    \small
    \centering
    \begin{tabular}{lp{12cm}}
    \toprule
        Question & What Caribbean country has a GDP eflator change rate of 2.32? \\ \midrule
        Subgraph ID (RoG) & location.location.containedby \\
        & common.topic.notable\_types \\
        & location.location.contains \\ \midrule
        Subgraph ID (GSR) & location.location.containedby \\
        & location.location.contains\\
        & location.statistical\_region.places\_exported\_to -- location.statistical\_region.places\_imported\_from \\ \midrule
        Retrived Subgraph & Caribbean $\leftarrow$ location.location.containedby $\leftarrow$ Puerto Rico \\
        & Caribbean $\leftarrow$ location.location.containedby $\leftarrow$ Barbados \\
        & Caribbean $\leftarrow$ location.location.containedby $\leftarrow$ Grace University, main campus \\
        & ... \\
        & Caribbean $\rightarrow$ location.location.contains $\rightarrow$ Puerto Rico \\
        & Caribbean $\rightarrow$ location.location.contains $\rightarrow$ Barbados \\
        & Caribbean $\rightarrow$ location.location.contains $\rightarrow$ Grace University, main campus \\
        & ... \\
        \midrule
        Reference Answer(s) & Puerto Rico \\
        \midrule
        Predicted Answers & Barbados \\
    \bottomrule
    \end{tabular}
    \caption{One error example due to the missing of constraint information sampled from CWQ test set. This is a common issue among sr-based KGQA works training with weakly supervised data.}
    \label{tab:constraint_example}
\end{table*}

In this section, we perform several case studies to better understand the methods and the pipeline.

\subsection{Subgraph ID Generation}
Table~\ref{tab:subgraphid_case} shows some subgraph ID generation sampled from WebQSP and CWQ test set. The first 2 is from WebQSP and the latter 2 is from CWQ. We can observe that GSR trained with raw data usually generate loop path which is helpless to the question and will bring unreliable recall (e.g., all person with same gender as topic entity will be retrieved in first example).

\subsection{Success Case}
Table~\ref{tab:case_success} shows a success example sampled from the CWQ dataset. Specially, when represent retrieved subgraph as triple sets, we only present repeated triples once. For example, (Lou Seal, sports.mascot.team, San Francisco Giants) appears 6 times when represent retrieved subgraph as paths (first 6 paths), but only appear once when represent retrieved subgraph as triples (first triple).

\subsection{Error Analysis}
In this part, we analyse when our proposed method fail in KGQA task. 
\label{app:limitation_constraint}
SR-based KGQA methods typically fail in following constraint in complex questions. Table~\ref{tab:constraint_example} shows an example that our proposed GSR model fails in this case. The question has a constraint of answer entity, which is `has a GDP eflator change rate of 2.32'. However, since in building training samples, we only care about shortest path between topic entity and answer entity, which means that the GDP information will not be considered in training (if this question is in training set) and in inference our model will miss such information. To this end, the retrieved subgraph do have the answer entity, but not enough for the reader to answer the question.

\end{document}